%% file: sample-sigconf.tex
\documentclass[sigconf]{acmart}
\usepackage{multirow}
\usepackage{bm}
\usepackage{physics}
\usepackage{balance}
\usepackage{amsmath}
\usepackage{makecell}
\newtheorem{prop}{Proposition}[section]
\newtheorem{assump}{Assumption}[section]
\usepackage[linesnumbered,boxed,ruled,vlined]{algorithm2e}
\newcommand\hemdash{\hbox{---}\kern-.5em---}

\newcommand{\eat}[1]{}
\usepackage[noabbrev,capitalize]{cleveref}

\copyrightyear{2021}
\acmYear{2021}
\acmConference[WWW '21]{Proceedings of the Web Conference 2021}{April 19--23, 2021}{Ljubljana, Slovenia}
\acmBooktitle{Proceedings of the Web Conference 2021 (WWW '21), April 19--23, 2021, Ljubljana, Slovenia}
\acmPrice{}
\acmDOI{10.1145/3442381.3449899}
\acmISBN{978-1-4503-8312-7/21/04}

\begin{document}
\title{Mask-GVAE: Blind Denoising Graphs via Partition}

\author{Jia Li}
\affiliation{%
  \institution{The Chinese University of Hong Kong}
}
\email{lijia@se.cuhk.edu.hk}

\author{Mengzhou Liu}
\affiliation{%
  \institution{The Chinese University of Hong Kong}
}
\email{mzliu@se.cuhk.edu.hk}

\author{Honglei Zhang}
\affiliation{%
  \institution{Tianjin Univerisity}
}
\email{zhanghl1996@tju.edu.cn}

\author{Pengyun Wang}
\affiliation{%
  \institution{Huawei Noah's Ark Lab}
}
\email{wangpengyun@huawei.com}

\author{Yong Wen}
\affiliation{%
  \institution{Huawei Noah's Ark Lab}
}
\email{wenyong4@huawei.com}

\author{Lujia Pan}
\affiliation{%
  \institution{Huawei Noah's Ark Lab}
}
\email{panlujia@huawei.com}

\author{Hong Cheng}
\affiliation{%
  \institution{The Chinese University of Hong Kong}
}
\email{hcheng@se.cuhk.edu.hk}

\renewcommand{\shortauthors}{J. Li et al.}

\begin{abstract}
We present Mask-GVAE, a variational generative model for blind denoising large discrete graphs, in which "blind denoising" means we don't require any supervision from clean graphs. We focus on recovering graph structures via deleting irrelevant edges and adding missing edges, which has many applications in real-world scenarios, for example, enhancing the quality of connections in a co-authorship network. Mask-GVAE makes use of the robustness in low eigenvectors of graph Laplacian  against random noise and decomposes the input graph into several stable clusters. It then harnesses the huge computations by decoding probabilistic smoothed subgraphs in a variational manner. On a wide variety of benchmarks, Mask-GVAE outperforms competing approaches by a significant margin on PSNR and WL similarity.

\end{abstract}



\maketitle

\input{samplebody-conf}

\appendix
\input{appendices}

\input{bibil_1.bbl}

\bibliographystyle{ACM-Reference-Format}
\bibliography{bibil_1}

\end{document}

%% file: samplebody-conf.tex
\section{Introduction}
\label{headings}

Recently, graph learning models \cite{kipf2017semi,pmlr-v97-wu19e} have achieved remarkable progress in many graph related tasks. Compared with other machine learning models that build on i.i.d.\ assumption, graph learning models require as input a more sophisticated graph structure. Meanwhile, how to construct a graph structure from raw data is still an open problem.  For instance, in a social network the common practice to construct graph structures is based on observed friendships between users \cite{jiawww19}; in a protein-protein interaction (PPI) network the common practice is based on truncated spatial closeness between proteins \cite{dobson2003distinguishing}.  However, these natural choices of graphs may not necessarily describe well the intrinsic relationships between the node attributes in the data \cite{dong2016learning,pang2017graph}, e.g., observed friendship does not indicate true social relationship in a social network \cite{zhang2018link}, truncated spatial closeness may incorporate noisy interactions and miss true interactions between proteins in a PPI network \cite{wang2018network}. In this work, we assume we are given a degraded graph structure, e.g., having missing/irrelevant edges. \emph{We aim to recover a graph structure that removes irrelevant edges and adds missing edges.} As in practice noisy-clean graph pairs are rare \cite{majumdar2018blind}, we propose to de-noise the input noisy graph without any supervision from its clean counterpart, which is referred to as \emph{blind graph denoising}.

Latent generative models such as variational autoencoders (VAEs) \cite{kingma2013auto} have shown impressive performance on denoising images \cite{im2017denoising} and speech \cite{chorowski2019unsupervised}.  As images or speech exist in a continuous space, it is easy to utilize gradient descent weapons to power the denoising process. On the contrary, our problem setting involves a large discrete structure, thus it is challenging to generalize the current deep generative models to our problem setting. 

\begin{figure*}
\begin{center}
\includegraphics [width=0.7\textwidth]{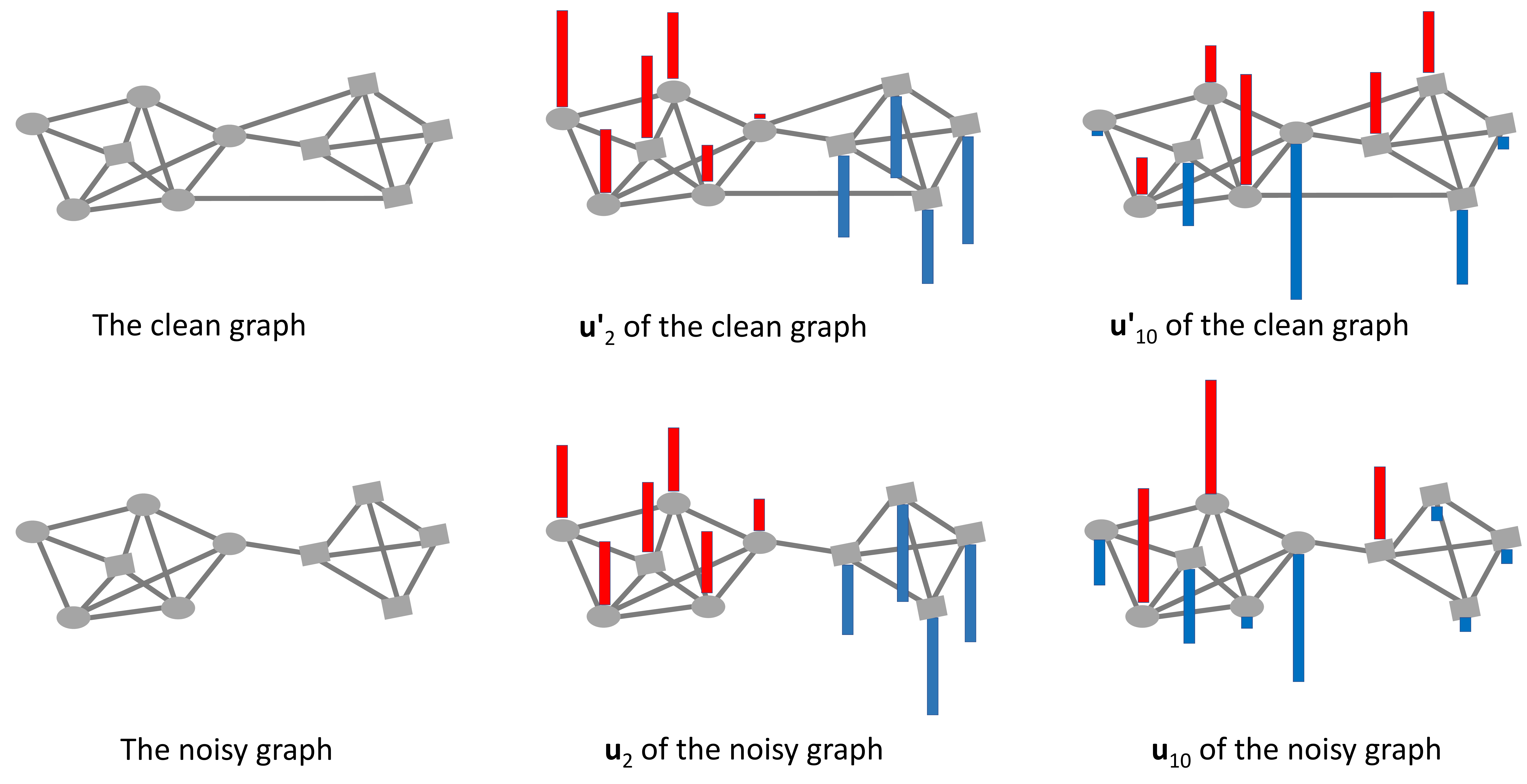}
\end{center}

\caption{Graph Laplacian eigenvectors of a graph and noisy one. $\mathbf{u}_2$ (the eigenvector w.r.t.\ the second smallest eigenvalue) is stable against random noise while $\mathbf{u}_{10}$ (the eigenvector w.r.t.\ the largest eigenvalue) fluctuates. }
\label{fig.ex}
\vspace{-0.3cm}
\end{figure*}

We present Mask-GVAE, the first variational generative model for blind denoising large graphs. A key insight of Mask-GVAE is that graph Laplacian eigenvectors associated with small eigenvalues (low eigenvectors) are stable against random edge perturbations if there is a significant cluster structure  \cite{karrer2008robustness,peng2002aver,eldridge2017unperturbed}. From another viewpoint, \cite{huang2009spectral} finds the stability of low eigenvectors lays the foundation of robustness of spectral clustering \cite{von2007tutorial}. Likewise, many real-world graphs do hold these distinct substructures, e.g., PPI network \cite{ahn2010link}, social network \cite{jiawww19}, and co-authorship network \cite{zhou2009graph}. We show an illustrative example in Figure \ref{fig.ex}. In Mask-GVAE, we first use graph neural networks (GNNs) \cite{kipf2017semi,li2020heatts} and \emph{normalized cut} \cite{shi2000normalized} for the fast estimation of low eigenvectors, i.e., cluster mask. We then encode the latent variables and generate probabilistic smoothed subgraphs, conditioned on the cluster mask and latent variables. A discrete graph is then sampled upon the denoised subgraphs to circumvent the non-differentiability problem \cite{simonovsky2018graphvae}.  

An important requirement of Mask-GVAE is the ability to fast estimate stable cluster masks.  While there are some neural networks proposed for computing cluster masks  \cite{cavallari2017learning,shaham2018spectralnet,chen2017supervised,nazi2019gap}, they either rely on other outsourcing tools such as K-means \cite{shaham2018spectralnet}, or require as input supervised information \cite{chen2017supervised}. We propose an end-to-end neural network to encode cluster mask in an unsupervised manner. It differs from the above methods as (1) it uses GNN for fast computation of low eigenvectors, as GNN quickly shrinks high eigenvectors and keeps low eigenvectors \cite{pmlr-v97-wu19e,li2020heatts}; (2)  it generalizes \emph{normalized cut} \cite{shi2000normalized} to work as the loss function, since the optimal solution of spectral relaxed \emph{normalized cut} coincides with the low eigenvectors  of normalized Laplacian \cite{von2007tutorial}.

Another challenge for Mask-GVAE is how to incorporate hard domain-specific constraints, i.e., cluster mask, into variational graph generation.  In the literature, GVAE \cite{kusner2017grammar} constructs parse trees based on the input graph and then uses Recurrent Neural Networks (RNNs) to encode to and decode from these parse trees.  It utilizes the binary \emph{mask} to delete invalid vectors. NeVAE \cite{samanta2019nevae} and CGVAE \cite{liu2018constrained} both leverage GNNs to generate graphs which match the statistics of the original data. They make use of a similar \emph{mask} mechanism to forbid edges that violate syntactical constraint. Our model also uses the \emph{mask} mechanism to generate cluster-aware graphs. 

Our contributions are summarized as follows.
\begin{itemize}
\item We study the blind graph denoising problem. Compared with the state-of-the-art methods that mainly focus on graphs with limited sizes, our solution is the first one that does not rely on explicit eigen-decomposition and can be applied to large graphs.

\item We present Mask-GVAE, the first variational generative model for blind denoising large graphs. Mask-GVAE achieves superior denoising performance to all competitors.

\item We theoretically prove low eigenvectors of graph Laplacian are stable against random edge perturbations if there is a significant cluster structure, which lays the foundation of many subgraph/cluster based denoising methods.

\item We evaluate Mask-GVAE on five benchmark data sets. Our method outperforms competing methods by a large margin on peak signal-to-noise ratio (PSNR) and Weisfeiler-Lehman (WL) similarity. 

\end{itemize}

The remainder of this paper is organized as follows.  Section \ref{def} gives the problem definition. Section \ref{alt} describes our methodology.  We provide a systematic theoretical study of the stability of low eigenvectors in Section \ref{sec.theory}.  We report the experimental results in Section \ref{sec.exp} and discuss related work in Section \ref{sec.related}.  Finally, Section \ref{sec.con} concludes the paper.

\section{Problem definition}\label{def}

Consider a graph $G = (V, A, X)$ where $V=\{v_1, v_2, \ldots, v_N\}$ is the set of nodes.  We use an $N\times N$ adjacency matrix $A$ to describe the connections between nodes in $V$.  $A_{ij}\in \{0, 1\}$ represents whether there is an undirected edge between nodes $v_i$ and $v_j$.  We use $X=\{x_1, x_2, \ldots, x_N\}$ to denote the attribute values of nodes in $V$, where $x_i \in \mathbb{R}^{d}$ is a $d$-dimensional vector.

We assume a small portion of the given graph structure $A$ is degraded due to noise, incomplete data preprocessing, etc. The corruptions are two-fold: (1) missing edges, e.g., missing friendship links among users in a social network, and (2) irrelevant edges, e.g., incorrect interactions among proteins in a protein-protein interaction network. The problem of \emph{\textbf{graph denoising}} is thus defined to recover a graph $\hat{G} = (V, \hat{A}, X)$ from the given one $G = (V, A, X)$. In this work, we don't consider noisy nodes and features, leaving this for future work.

Graph Laplacian regularization has been widely used as a signal prior in denoising tasks \cite{dong2016learning, pang2017graph}. Given the adjacency matrix $A$ and the degree matrix $D$ with $D_{ii} = \sum_j A_{ij}$, the graph Laplacian matrix is defined as $L = D - A$. Recall $L$ is a positive semidefinite matrix, $X_{:,k}^{\top}LX_{:,k} = \frac{1}{2}\sum_{i,j=1}^NA_{ij}(X_{ik} - X_{jk})^2$ measures the sum of pairwise distances between nodes in $G$, and $X_{:,k}$ is the $k$-th column vector of $X$. In this work, we consider the recovered structure $\hat{A}$ should be \emph{coherent} with respect to the features $X$.  In this context, the graph denoising problem has the following objective function:

\begin{equation}
  \arg\min{} \ \ \mathcal{L} = \sum_{i<j}|A_{ij} - \hat{A}_{ij}| + \omega\Tr(X^{\top}\hat{L}X),\\
\label{equ.ctotal}
\end{equation}
where the first term is a fidelity term ensuring the recovered structure $\hat{A}$ does not deviate from the observation $A$ too much, and the second term is the graph Laplacian regularizer. $\hat{L}$ is the graph Laplacian matrix for $\hat{A}$. $\omega \geq 0$ is a weighting parameter. $\Tr(\cdot)$ is defined as the sum of elements on the main diagonal of a given square matrix. In this work, we focus on 
\emph{\textbf{blind graph denoising}}, i.e., we are unaware of the clean graph  and the only knowledge we have is the observed graph $G = (V, A, X)$. 

Our problem formulation is applicable to many real applications as illustrated below.
\paragraph{Application 1} In many graph-based tasks where graph structures are not readily available, one needs to construct a graph structure first \cite{pang2017graph,zeng2019deep, hu2019feature}. As an instance, the k-nearest neighbors algorithm (k-NN) is a popular graph construction method in image segmentation tasks \cite{felzenszwalb2004efficient}. One property of a graph constructed by k-NN is that each pixel (node) has a fixed number of nearest neighbors, which inevitably introduces irrelevant and missing edges. Thus, graph denoising can be used to get a denoised graph $\hat{A}$ by admitting small modifications with the observed signal prior $A$.
\paragraph{Application 2} Consider a knowledge graph (KG) where nodes represent entities and edges encode predicates. It is safe to assume there are some missing connections and irrelevant ones, as KGs are known for their incompleteness \cite{zhang2019few}.  In this context, the task of graph denoising is to modify the observed KG such that its quality can be enhanced.

\paragraph{Application 3} Consider a co-authorship network as another example where a node represents an author and an edge represents the co-authorship relation between two authors.  Due to the data quality issue in the input bibliographic data (e.g., name ambiguity) \cite{kim2016distortive}, the constructed co-authorship network may contain noise in the form of irrelevant and missing edges.  In this scenario, our graph denoising method can be applied to the co-authorship network to remove the noise.

We then contrast the difference between graph denoising and other related areas including adversarial learning and link prediction below.
\paragraph{graph denoising vs. adversarial learning}
An adversarial learning method considers a specific task, e.g., node classification \cite{zugner2018adversarial}. In this regard, there is a loss function (e.g., cross-entropy for node classification) guiding the attack and defense. Differently, we consider the general graph denoising without any task-specific loss. We thus don't compare with the literature of adversarial learning in this work.
\paragraph{graph denoising vs. link prediction}
While link prediction is termed as predicting whether two nodes in a network are likely to have a link \cite{liben2007link}, currently most methods \cite{liben2007link,zhang2018link} focus on inferring missing links from an observed network. Differently, graph denoising considers the observed network consists of both missing links and irrelevant links. Moreover, while link prediction can take advantage of a separate trustworthy training data to learn the model, our blind setting means our model needs to denoise the structures based on the noisy input itself.

\section{Methodology}\label{alt}

Motivated by the recent advancements of  discrete VAEs \cite{lorberbom2019direct} and denoising VAEs \cite{im2017denoising, chorowski2019unsupervised} on images and speech, we propose to extend VAEs to the problem of denoising large discrete graphs. However, current graph VAEs still suffer from the scalability issues \cite{salha2019degeneracy,salha2020fastgae} and cannot be generalized to large-scale graph settings. As another issue, current VAEs suffer from a particular local optimum known as component collapsing \cite{kingma2016improved}, meaning a good optimization of prior term results in a bad reconstruction term; if we directly attaching  Eq.1 and the loss function of VAE, the situation would be worse. 
In this work, we present Mask-GVAE, which first decomposes a large graph into stable subgraphs and then generates smoothed subgraphs in a variational manner.
Specifically, Mask-GVAE consists of two stages, one computing the cluster mask and the other generating the denoised graph.

\subsection{Cluster mask}\label{acdm}
A cluster mask $C$ encodes the stable low eigenvectors of graph Laplacian. Specifically, $C \in \{0, 1\}^{N\times K}$ is a binary mask, where $K$ is the number of clusters, $C_{ik} = 1$ represents node $i$ belongs to cluster $k$ and $C_{ik} = 0$ otherwise. 

\paragraph{The loss function}\label{dem}
The definition of graph cut is:

\begin{equation}
   \frac{1}{K}\sum_{k}cut(V_k,\overline{V_k}),
\label{equ.H1}
\end{equation}
where $V_k$ is the node set assigned to cluster $k$, $\overline{V_k} = V\setminus V_k$, $cut(V_k,\overline{V_k}) = \sum_{i \in V_k, j \in \overline{V_k}} A_{ij}$ and it calculates the number of edges with one end point inside cluster $V_k$ and the other in the rest of the graph.  Taking $C$ into consideration, the graph cut can be re-written as:

\begin{equation}
    \frac{1}{K}\sum_{k}(C_{:,k}^\top DC_{:,k} - C_{:,k}^\top AC_{:,k})
  =\frac{1}{K}\Tr(C^\top LC),
\label{equ.H3}
\end{equation}
 in which $D$ and $L$ are the degree and Laplacian matrices respectively. $C_{:,k}^\top DC_{:,k}$ stands for the number of edges with at least one end point in $V_k$ and $C_{:,k}^\top AC_{:,k}$ counts the number of edges with both end points in cluster $V_k$. The \emph{normalized cut} \cite{shi2000normalized,nazi2019gap} thus becomes:
 \begin{equation}
   \frac{1}{K}\Tr((C^\top LC)\oslash(C^\top DC)),
\end{equation}
where $\oslash$ is element-wise division.
 Note an explicit constraint is that $C^\top C$ is a diagonal matrix, we thus apply a penalization term \cite{li2020adversarial}, which results in a differentiable unsupervised loss function:
\begin{equation}
  \mathcal{L}_u = \frac{1}{K}\Tr((C^\top LC)\oslash(C^\top DC)) + \varphi \big|\big|\ \frac{K}{N}C^\top C - I_K\ \big|\big|_F^2,
\label{equ.Hu}
\end{equation}
where $\big|\big|\cdot\big|\big|_F$ represents the Frobenius norm of a matrix.

\paragraph{The network architecture}\label{ssc}
  Our architecture is similar to \cite{nazi2019gap,li2020adversarial} , which has two main parts: (1) node embedding, and (2) cluster assignment. In the first part, we leverage two-layer graph neural networks, e.g., GCN \cite{kipf2017semi}, Heatts \citep{li2020heatts}, to get two-hop neighborhood-aware node representations, thus those nodes that are densely connected and have similar attributes can be represented with similar node embeddings. In the second part, based on the node embeddings, we use two-layer perceptrons and softmax function to assign similar nodes to the same cluster. The output of this neural network structure is the cluster mask $C$, which is trained by minimizing the unsupervised loss $\mathcal{L}_u$ in Eq.\ \ref{equ.Hu}.

\begin{figure*}
\begin{center}
\includegraphics [width=0.8\textwidth]{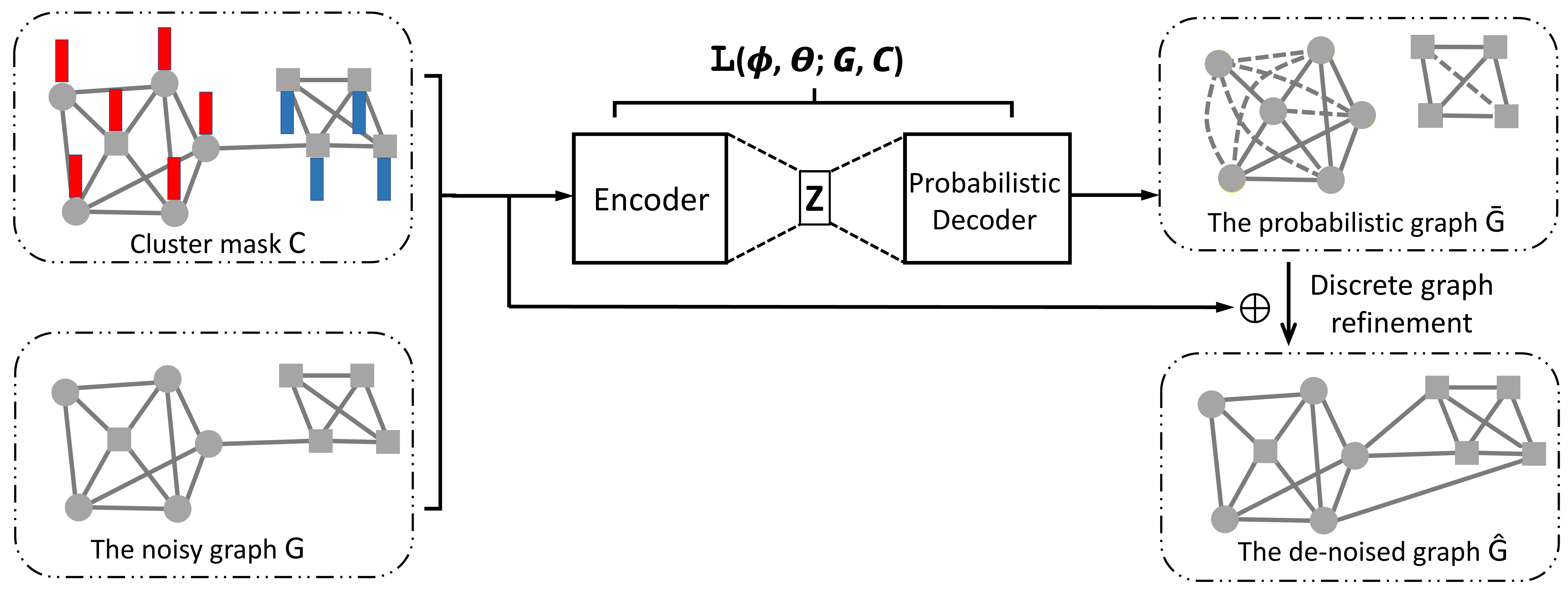}
\end{center}

\caption{The proposed  Mask-GVAE. It first utilizes the cluster mask $C$ and the noisy graph $G$ to learn the probabilistic de-noised graph $\bar{G}$. The learning process is conditioned on $C$ as cluster results are robust against random noise (Proposition \ref{thm:prop}). With the optimum of the learning process, it then draws discrete graph $\hat{G}$ at test time to meet the fidelity constraint.}
\label{fig.sape}
\end{figure*}

\subsection{Denoised graph generation}\label{DGG}
In this subsection, we describe our method (illustrated in Figure \ref{fig.sape}) which produces a denoised graph $\hat{A}$ that is conditioned on the cluster mask $C$ and meets the fidelity constraint.  As discussed, we use a latent variable model parameterized by neural networks to generate the graph $\hat{G}$. Specifically, we focus on learning a parameterized distribution over the graph $G$ and the cluster mask $C$ as follows:
\begin{equation}
  P(\hat{A}|G,C) = \int q_{\phi}(Z|G,C)p_{\theta}(\hat{A}|G,C,Z)dZ,
\label{equ.ge}
\end{equation}
where $q_{\phi}(Z|G,C)$ is the encoder and $p_{\theta}(\hat{A}|G,C,Z)$ is the decoder.  While the encoder is straightforward and we can use the corresponding encoders in existing work \cite{kipf2016variational}, the decoder is hard due to the following two factors:
\begin{itemize}
\item  \emph{Discrete decoding}: Generating discrete graphs is challenging.
\item  \emph{Cluster awareness}: Existing graph generation methods cannot explicitly incorporate cluster information.
\end{itemize}

To satisfy the \emph{discrete decoding}, we decouple the decoder into two steps: (1) probabilistic graph decoder, which produces a probabilistic graph $\bar{A}$ in training, and (2) discrete graph refinement, in which we sample a discrete graph $\hat{A}$ in testing based on the prior knowledge $G$ and $\bar{A}$. To address the \emph{cluster awareness}, we propose to directly incorporate cluster information into the learning process by using the \emph{mask} \cite{kusner2017grammar} mechanism, which powers the model with the ability to generate denoised graph with smoothed subgraphs.

\paragraph{Graph encoder}
We follow VGAE \cite{kipf2016variational} by using the mean field approximation to define the variational family:
\begin{equation}
  q_{\phi}(Z|G,C) = \prod_{i=1}^Nq_{\phi_i}(z_i|A,X),
\label{equ.enc}
\end{equation}
where $q_{\phi_i}(z_i|A,X)$ is the predefined prior distribution, namely, isotropic Gaussian with diagonal covariance.
The parameters for the variational marginals $q_{\phi_i}(z_i|A,X)$ are specified by a two-layer GNN \cite{kipf2017semi,li2020heatts}:
\begin{equation}
 \mu,\sigma = \text{GNN}_{\phi}(A,X),
\label{equ.H9}
\end{equation}
where $\mu$ and $\sigma$ are the vector of means and standard deviations for the variational marginals $\{q_{\phi_i}(z_i|A,X)\}_{i=1}^N$. $\phi =\{\phi_i\}_{i=1}^N$ is the parameter set for encoding.

\paragraph{Probabilistic graph decoder}

We first compute the edge probability:
\begin{equation}
p_{\theta}(\bar{A}_{ij}|G,C,Z)= \bar{A}_{ij}= \text{sigmoid} (W_{a1}\text{ReLU}(W_{a2}E_{ij})),
\end{equation}
where $E_{ij} = [Z_i|X_i]\odot[Z_j|X_j]$, $Z_i|X_i$ is a concatenation of $Z_i$ and $X_i$, $\odot$ represents element-wise multiplication. Intuitively we compute the edge probability by two-layer perceptrons and sigmoid function. 
\paragraph{The loss function}
We approximate $p(\bar{A}|G,C,Z)$ by:
\begin{align}
\begin{split}
 p(\bar{A}|G,C,Z) = \prod_{(CC^\top)_{ij} = 1}\bar{A}_{ij}  \prod_{(CC^\top)_{ij} = 0}(1-\bar{A}_{ij}),
\end{split}
\end{align}
where $(CC^\top)_{ij} = 1$ denotes node $i$ and node $j$ are in the same cluster and $(CC^\top)_{ij} = 0$ otherwise. When the number of clusters $K$ is large, it is beneficial to re-scale both $CC^\top_{ij} = 1$ and $CC^\top_{ij} = 0$ within $A_{ij} = 1$ . 
\begin{equation}
  S = CC^\top\odot A,
\label{HS}
\end{equation}
where $S_{ij} = 1$ denotes node $i$ and node $j$ are adjacent and in the same cluster w.r.t.\ the cluster mask $C$. With $S$, $(A - S)$ can be used to denote the inter-cluster connections. 

The \textbf{overall loss} for Mask-GVAE is:
\begin{equation}
\begin{split}
\mathcal{L}(\phi,\theta;G,C) = \mathcal{L}_{prior} -\mathbb{E}_{q_\phi(Z/G,C)}(\log p(\bar{A}|G,C,Z)),
\end{split}
\label{overall}
\end{equation}
where $\mathcal{L}_{prior} = \text{KL}(q(Z|G,C)||p(Z))$ is the prior loss with $p(Z) = \prod_{i=1}^NP(z_i)=\prod_{i=1}^N\mathcal{N}(z_i|0,I)$. Intuitively, Eq.\ \ref{overall} takes into consideration both the objective function in Eq.\ \ref{equ.ctotal} and cluster structures.

\paragraph{Discrete graph refinement}
At testing time, we draw discrete samples from the probabilistic graph $\bar{G} = (V, \bar{A},X)$. To ensure the de-noised graph $\hat{A}$ does not deviate from the observation $A$ too much, we set a budget $\Delta$ to measure their difference.   
We split the budget into two parts, $\Delta/2$ for adding edges, and $\Delta/2$ for deleting edges. The refinement strategy is simply sampling without replacement from the formulated Category distributions. 

To delete edges, we restrict ourselves to existing inter-cluster connections $(A - S)_{ij} = 1$.
\begin{equation*}
 E_{ij}^- = \frac{\exp{(1 - \bar{A}_{ij})}}{\sum \exp{( 1 - \bar{A}_{ij})}}\ \ \ \ if\ \ \ \ (A - S)_{ij}=1.
\end{equation*}
We sample without replacement $\Delta/2$ edges according to their probabilities $E_{ij}^-$. The sampled edge set is then deleted from the original graph structure $A$. To add edges, we are allowed to add connections for intra-cluster nodes $(CC^\top - A)_{ij} = 1$.
\begin{equation*}
 E_{ij}^+ = \frac{\exp{ \bar{A}_{ij}}}{\sum \exp{ \bar{A}_{ij}}}\ \ \ \ if\ \ \ \ (CC^\top - A)_{ij}=1.
\end{equation*}
We sample $\Delta/2$ edges according to their probabilities $E_{ij}^+$. The sampled edge set is added into the original graph structure $A$.  With this strategy, we can generate the final denoised graph structure $\hat{A}$.

\paragraph{Connection between Mask-GVAE and Eq.\ \ref{equ.ctotal}}
The first term of Eq.\ \ref{equ.ctotal} is used to ensure the proximity between the input graph $A$ and the denoised graph $\hat{A}$, which is achieved in Mask-GVAE by the reconstruction capacity of VAEs and the discrete budget in the sampling stage. Next, we illustrate the connection between the graph Laplacian term in Eq.\ \ref{equ.ctotal} and Mask-GVAE.  We can re-write the graph Laplacian regularization when $d = 1$ as:
\begin{equation}
 X^{\top}LX_{d = 1} = \frac{1}{2} (\underbrace{\sum_{S_{ij} = 1}(X_i - X_j)^2}_{intra-cluster} + \underbrace{\sum_{(A -S)_{ij} = 1}(X_i - X_j)^2}_{inter-cluster}),
 \label{e.an}
\end{equation}
By decomposing the  Laplacian regularization into these two terms, we focus on intra/inter-cluster connections when adding/deleting edges. As attributed graph clustering aims to discover groups of nodes in a graph such that the intra-group nodes are not only more similar but also more densely connected than the inter-group ones \cite{wang2015community}, this heuristic contributes to the overall optimization of the graph Laplacian term.

\subsection{The Proposed Training Method}\label{CGG}
Our solution consists of a cluster mask module $f(\cdot)$ and a denoised graph generation module $g(\cdot)$. As the error signals of the denoised graph generation module are obtained from the cluster mask module and the cluster mask module needs the denoised graphs as inputs for robust cluster results, we design an iterative framework, to alternate between minimizing the loss of both $g(\cdot)$ and $f(\cdot)$. We refer to Algorithm \ref{algo.set} for details of the training procedure.

We denote the parameters of the cluster mask module $f(\cdot)$ as $\mathcal{W}_{f}$ and the parameters of the denoised graph generation module $g(\cdot)$ as $\mathcal{W}_{g}$. At the beginning of Algorithm \ref{algo.set}, we use the cluster mask module $f(\cdot)$ to get the initial cluster results. We then exploit the denoised graph generator $g(\cdot)$ so as to get a denoised graph $\hat{G}$ (line 4). Based on that,  we utilize the idea of \emph{randomized smoothing} \cite{jia2020certified, cohen2019certified} and feed both $G$ and $\hat{G}$ into the cluster mask module $f(\cdot)$ to compute $\mathcal{L}_{u}$ (line 7). With the optimum of $f(\cdot)$, we get the robust cluster mask $C$ to power the learning process of $g(\cdot)$ (line 9).

\begin{algorithm}
  \caption{Training the model}
  \label{algo.set}
  \KwIn{$G$, $K$.}
  \KwOut{$\hat{G}$.}
  Initial: parameters $\mathcal{W}_{g}$,$\mathcal{W}_{f}$;

  \Repeat{deadline}{
    $C \leftarrow$ $f(G,K)$\;
	$\hat{G} \leftarrow$ $g(G,C)$\;
    $\mathcal{L}_{prior} \leftarrow \text{KL}(q(Z|G,C)||p(Z))$\;
	$\mathcal{L}(\phi,\theta;G,C)  \leftarrow  \mathcal{L}_{prior} -\mathbb{E}_{q_\phi(Z/G,C)}(\log p(\bar{A}|G,C,Z))$\;
	$\mathcal{L}_{u} \leftarrow f(\{G,\hat{G}\})$\;
    // Update parameters according to gradients

    $\mathcal{W}_{g} \leftarrow^+ -\bigtriangledown_{\mathcal{W}_{g}} \mathcal{L}(\phi,\theta;G,C)$\;
    $\mathcal{W}_{f} \leftarrow^+ -\bigtriangledown_{\mathcal{W}_{f}} \mathcal{L}_{u}$\;
  }
\end{algorithm}

\subsection{Complexity Analysis}
We analyze the computational complexity of our proposed method.  The intensive parts of Mask-GVAE contain the computation of Eq. \ref{equ.Hu} and Eq. \ref{overall}. Aside from this, the convolution operation takes $O(|E|dh)$ \cite{kipf2017semi} for one input graph instance with $|E|$ edges where $h$ is the number of feature maps of the weight matrix.

Regarding Eq. \ref{equ.Hu}, the core is to compute the matrices $(C^\top LC)$ and $(C^\top DC)$. Using sparse-dense matrix multiplications, the complexity is $O(|E|K^2)$. For Eq. \ref{overall}, the intensive part is the topology reconstruction term. As the optimization is conditioned on existing edges, the complexity is $O(|E|)$.  Thus, it leads to the complexity of $O(|E|K^2 +|E|dh )$ for one input graph instance.

We compare with VGAE \cite{kipf2016variational} and one of our baseline DOMINANT \cite{ding2019deep}, whose operations require several convolution layers with $O(|E|dh)$ and topology reconstruction term on all possible node pairs $O(N^2)$. Thus, the complexity of VGAE and DOMINANT is $O(N^2 +|E|dh)$. As observed in the experiments, usually we have $|E|K^2 \leq N^2$ for large graphs, e.g., Citeseer, Pubmed, Wiki, Reddit. When compared with NE \cite{wang2018network} and ND \cite{Soheil2013network}, our solution wins as both NE and ND rely on explicit eigen-decomposition.

\section{Assumption analysis}\label{sec.theory}
The stability of cluster mask under random noise is vital for Mask-GVAE, and that stability is dominated by the stability of low eigenvectors of graph Laplacian \cite{von2007tutorial}.
In this part, we initiate a systematic study of the stability of low eigenvectors, using the notion of average sensitivity \cite{varma2019average}, which is the expected size of the symmetric difference
of the output low eigenvectors before and after we randomly remove a few edges.
Let the clean graph be $G' = (V,A',X)$, the degree matrix be $D'$ and the clean Laplacian matrix be $L' = D' - A'$. We denote the $i$-th smallest eigenvalue of $L'$ as $\lambda'_i$, and the corresponding eigenvector as $\mathbf{u'}_i$. We consider the noisy graph $G = (V,A,X)$ is generated from the clean one $G'$ by the following procedure: for each edge $(i,j)$ in $G'$, remove it with probability $q$ independently. Following \cite{huang2009spectral}, we analyze the bi-partition of the graph data, as a larger number of clusters can be found by applying the bi-partition algorithm recursively.

\begin{assump}
     Let the node degrees of the clean graph be $\sum\limits_{i=1}^{N} d{'}_{i}^{2} = N^{\chi}$ and the number of edges be $m' = N^{\varphi}$, the following properties hold.
     \begin{enumerate}
\item $\exists \epsilon \geq 2, s.t., \frac{2\epsilon}{\epsilon - 1}\lambda'_2 < \lambda'_3$ and $\lambda'_3 \geq \max(\frac{25}{2}\epsilon q\lambda'_N, \frac{3}{2}\epsilon\log N)$.
\item $q \leq \frac{\lambda'_2}{ \kappa N^{\beta}}$ and $\beta \geq \frac{1}{2} \max(\chi, \varphi)$ with $\kappa > 1$.
\end{enumerate}     

    \label{thm:ass}
\end{assump}
Assumption \ref{thm:ass}.1 implies the graph has at most one outstanding sparse cut by the higher-order Cheeger inequality \cite{lee2014multiway,peng2002aver}. It has been discussed that the eigenspaces of Laplacian with such a large eigengap are stable against edge noise \cite{von2007tutorial}. Assumption \ref{thm:ass}.2 indicates the probability of edge noise is small w.r.t.\ the number of nodes. To better understand Assumption \ref{thm:ass}, let us consider the example in Figure \ref{fig.ex}. It is easy to check $\chi = 2.32$, $\varphi = 1.3$, $\lambda'_2 = 0.78$, $\lambda'_3 =3.28$ and $\lambda'_N =6.96$. Let $q = 0.01$, then we can get $\beta = 1.16$ and $\kappa = 5.4$.

\begin{table}
  \centering
  \caption{Statistics of graphs used in graph clustering}
  \label{tab:sqq}
  \begin{tabular}{ccccc}
    \toprule
    \textbf{Data}&\textbf{Nodes}&\textbf{Edges}&\textbf{Class}&\textbf{features}\\
    \midrule
	Citeseer&3,327&4,732&6&3,703\\
	Pubmed&19,717&44,338&3&500\\
	Wiki&2,405&17,981&17&4,973\\
  \bottomrule
\end{tabular}
\end{table}

Denote $\mathbb{E}  [\sin(\angle(\mathbf{u'}_2, \mathbf{u}_2))] $ as the expected $\sin(\angle(\cdot,\cdot))$ for the angle between $\mathbf{u'}_2$ and $\mathbf{u}_2$, then the following proposition holds.

\begin{prop}
     Under Assumption \ref{thm:ass},  $\mathbb{E}  [\sin(\angle(\mathbf{u'}_2, \mathbf{u}_2))] $ under random noise satisfies $\mathbb{E}  [\sin(\angle(\mathbf{u'}_2, \mathbf{u}_2))] \leq \frac{1}{\kappa}$.
    \label{thm:prop}
\end{prop}
Please refer to Appendix \ref{a.d} for the proof. For the example in Figure \ref{fig.ex}, we can check $\mathbb{E} [\sin(\angle(\mathbf{u'}_2, \mathbf{u}_2))] \approx 0.07 \leq 1/5.4$.

\section{Experiments}\label{sec.exp}
We first validate the effectiveness of our graph clustering algorithm. Then we evaluate Mask-GVAE on blind graph denoising tasks.
\subsection{Graph clustering}
\eat{
\begin{figure*}
\centering
\includegraphics [width=0.328\textwidth]{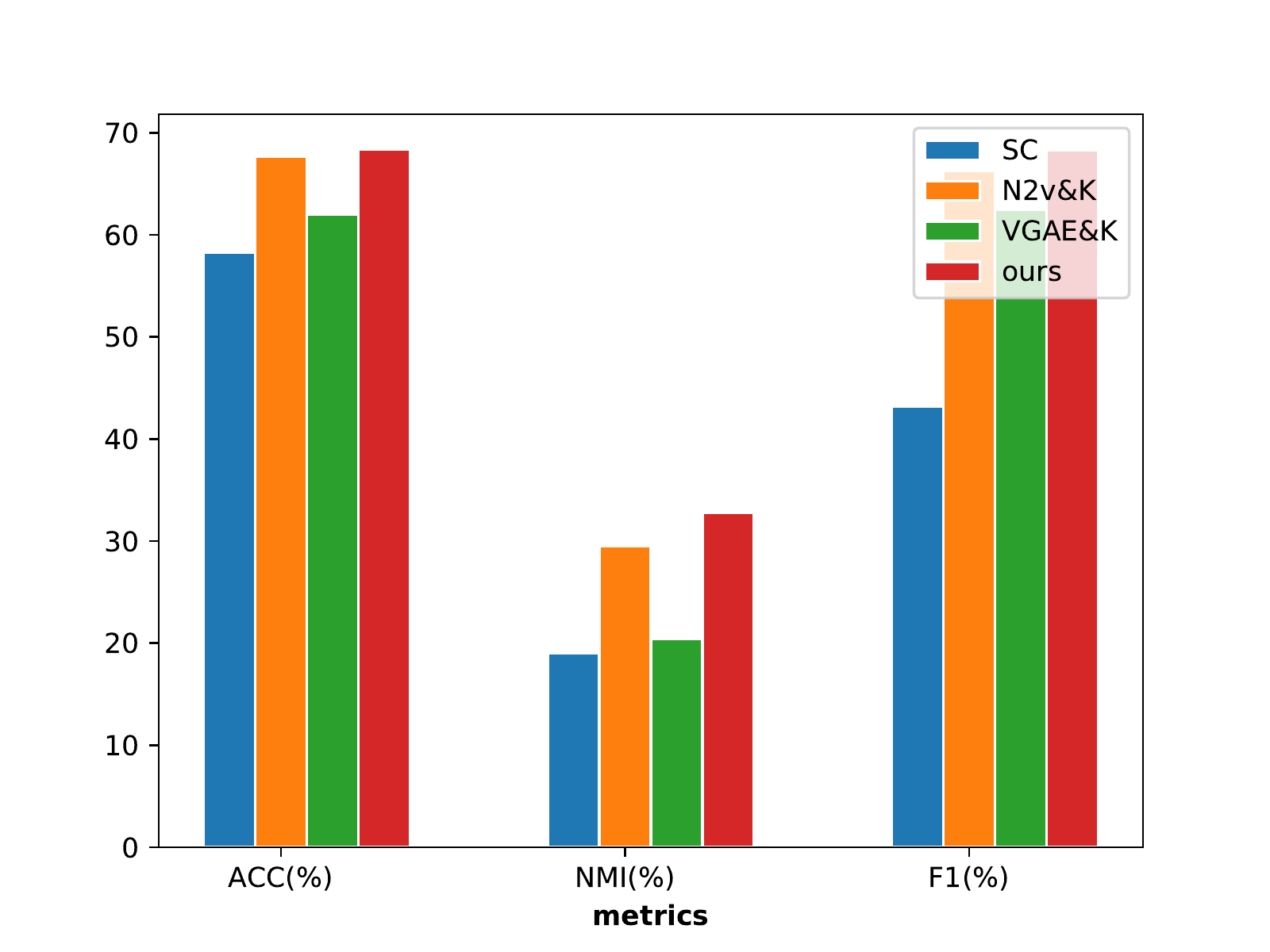}
\includegraphics [width=0.328\textwidth]{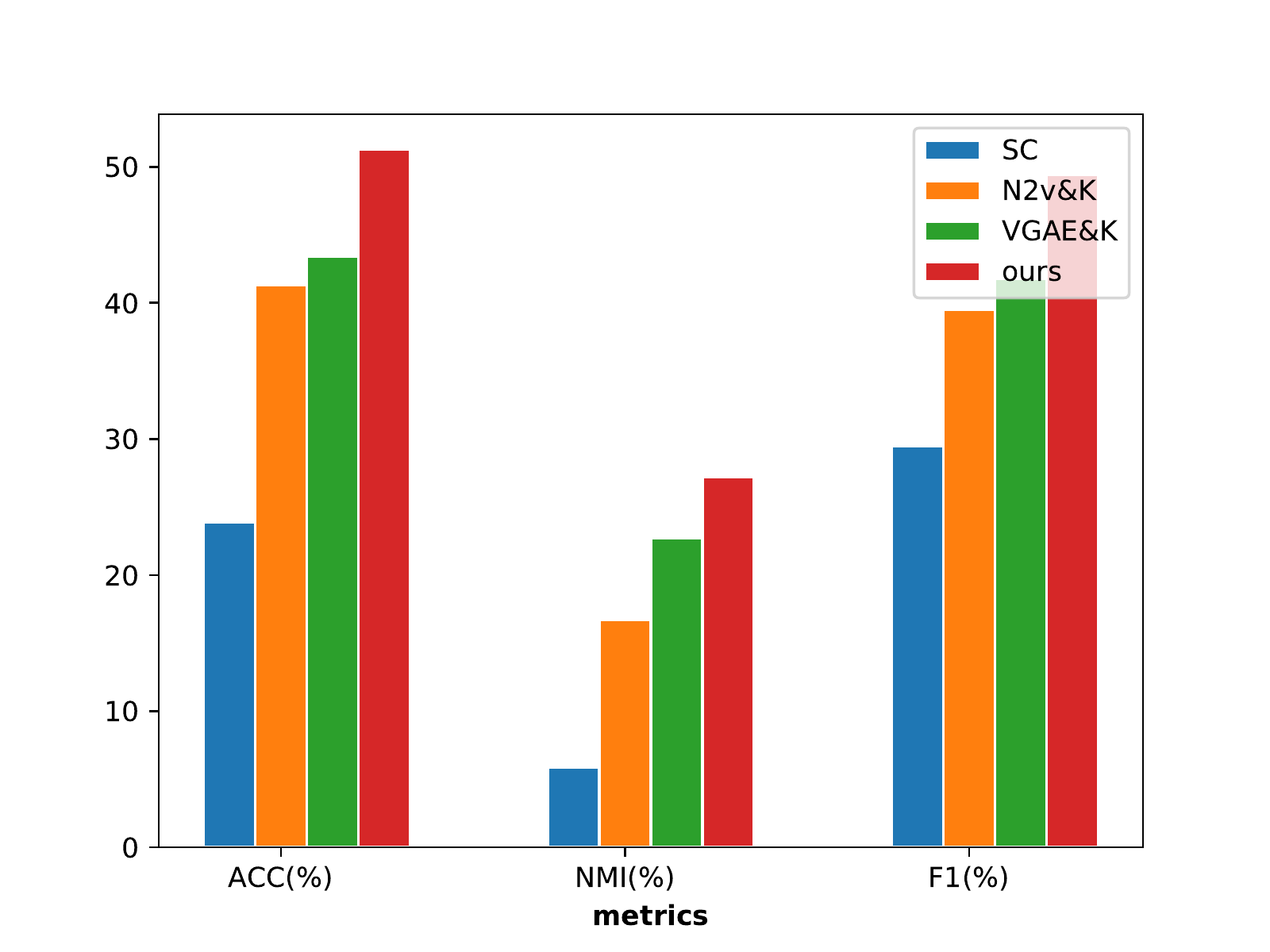}
\includegraphics [width=0.328\textwidth]{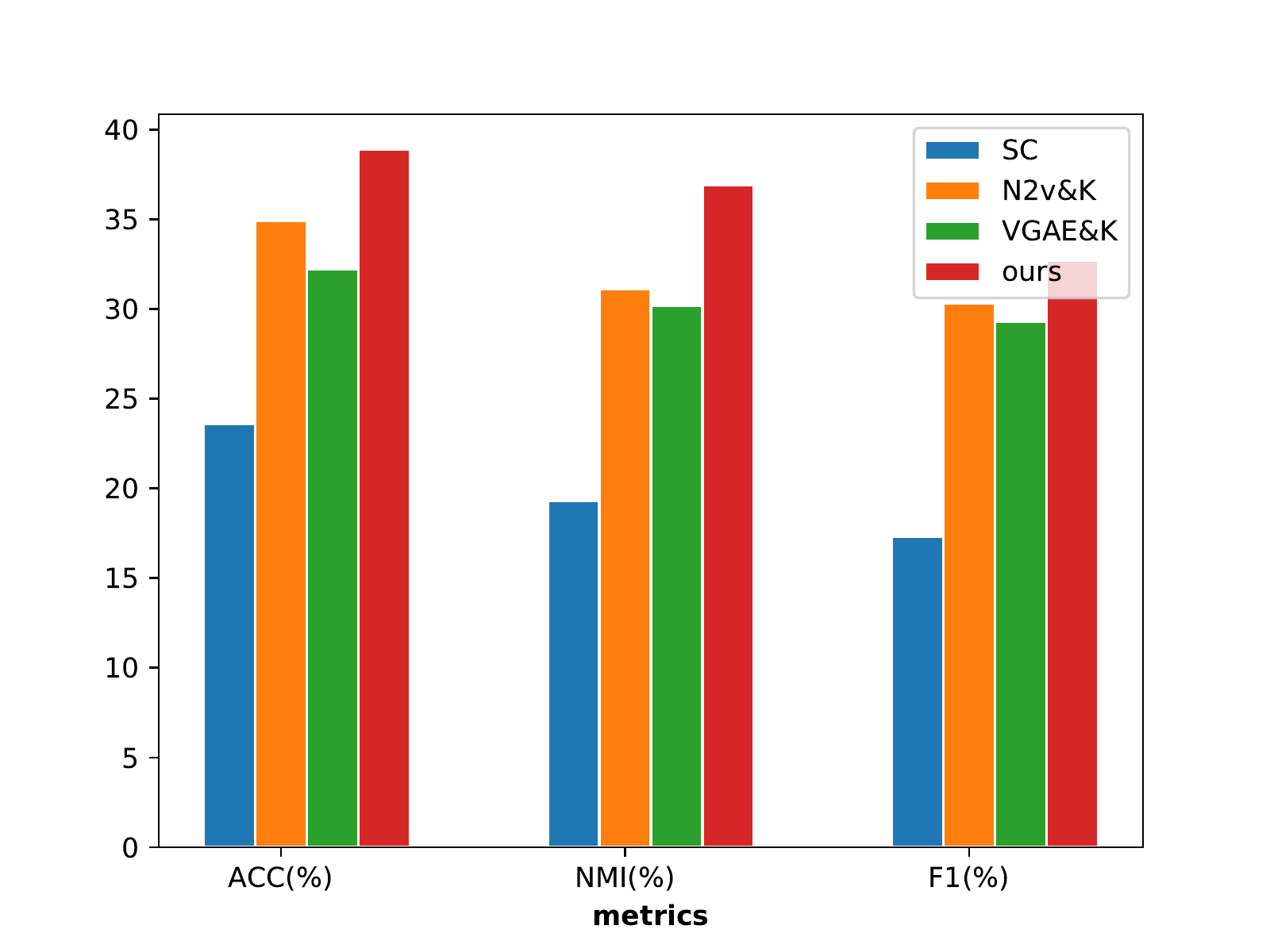}
\caption{Cluster performance comparison of different methods}
\label{loworder}
\end{figure*}
}
\begin{figure*}
\centering
\includegraphics [width=0.9\textwidth]{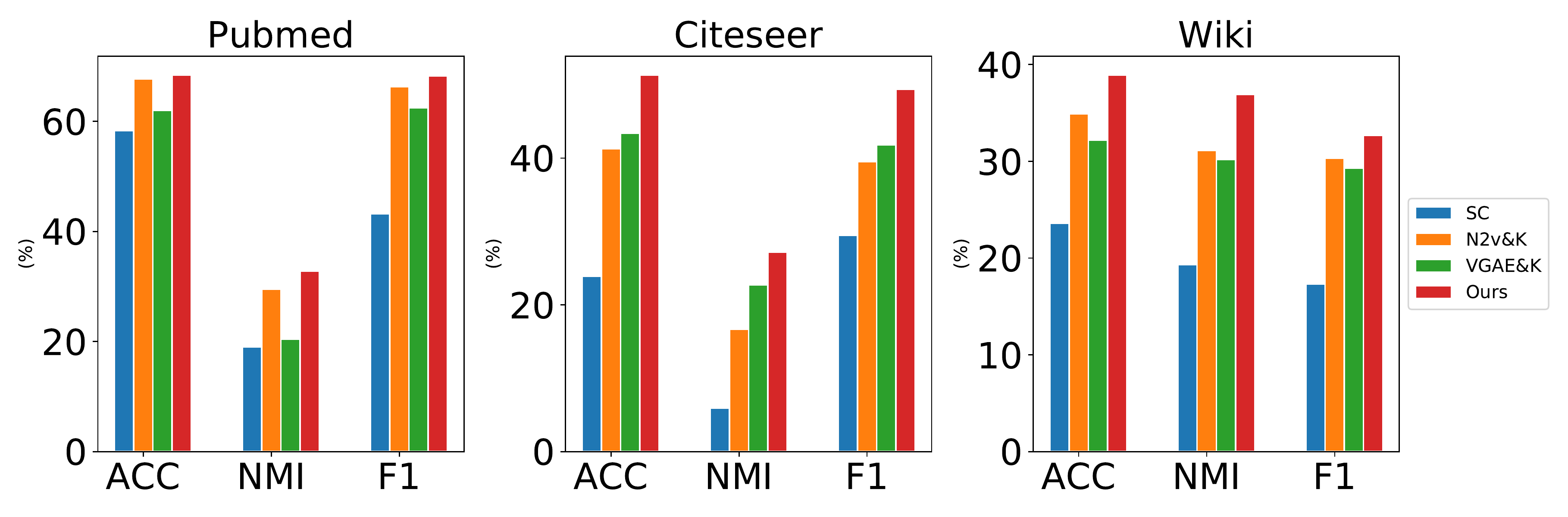}

\caption{Graph clustering performance comparison of different methods}
\label{gcluter}
\end{figure*}

\eat{
 \begin{table*}
  \caption{Cluster performance comparison of different methods}
  \scalebox{0.95}{
  \begin{tabular}{cccccccccc}
    \toprule
&\multicolumn{3}{c}{Pubmed}&\multicolumn{3}{c}{Citeseer}&\multicolumn{3}{c}{Wiki}\\
\midrule
        &ACC (\%) &NMI (\%) &F1 (\%) &ACC (\%) &NMI (\%) &F1 (\%) &ACC (\%) &NMI (\%) &F1 (\%)\\
    \midrule
    SC &58.3&19.0&43.2&23.9&5.9&29.5&23.6&19.3&17.3\\
	N2v\&K &67.7&29.5&66.3&41.3&16.7&39.5&34.9 &31.1&30.3\\
	VGAE\&K &62.0&20.4&62.5&43.4&22.7&41.8&32.2&30.2&29.3\\
	\midrule
	Ours &\textbf{68.4}&\bf 32.8 &\textbf{68.3}&\textbf{51.3}&\textbf{27.2}&\textbf{49.4}&\textbf{38.9}&\textbf{36.9}&\bf 32.7\\
  \bottomrule
\end{tabular}
}
 \label{gcluter}
\end{table*}
}
\paragraph{Data and baselines}
We use three benchmark data sets, i.e., Pubmed, Citeseer \cite{sen2008collective} and Wiki \cite{yang2015network}.  Statistics of the data sets can be found in Table \ref{tab:sqq}. As baselines, we compare against (1) Spectral Clustering (SC) \cite{von2007tutorial}, which only takes the node adjacency matrix as affinity matrix; (2) Node2vec \cite{grover2016node2vec} + Kmeans (N2v\&K), which first uses Node2vec to derive node embeddings and then utilizes K-means to generate cluster results; and (3) VGAE \cite{kipf2016variational} + Kmeans (VGAE\&K).

\paragraph{Setup}
For our method, we let the output dimension of the second-layer perceptrons equal to the number of clusters. We use the same network architecture through all the experiments. Our implementation is based on Tensorflow.  We train the model using full batch based Adam optimizer with exponential decay. We set $\varphi = 0.01$.  The output dimension of the first-layer GNN is 32. The best variants of GNN are chosen from GCN \citep{kipf2017semi} and Heatts \citep{li2020heatts}.  In addition, we use a dropout rate of 0.3 for all layers. The initial learning rate is 0.01. We leverage node class labels as the ground truth for graph clustering.
\paragraph{Results} The clustering accuracy (ACC), normalized mutual information (NMI) and macro F1 score (F1) are shown in Figure \ref{gcluter}. Our method outperforms the competitors on all data sets.  As ours does not rely on K-means to derive cluster memberships, this cluster performance indicates the effectiveness of our framework on graph clustering tasks.

\subsection{Graph denoising}

\paragraph{Baselines}
We use the following approaches as our baselines:
\begin{itemize}

\item DOMINANT \cite{ding2019deep}, a graph neural network that performs anomaly detection on attributed graph. It computes the degree of anomaly by the distance of recovered features/edges and the input features/edges. In this work, we consider the top anomaly edges as noisy edges and remove them from the input graphs. 

\item NE \cite{wang2018network}, a blind graph denoising algorithm that recovers the graph structure based on diffusion algorithms. It is designed to remove irrelevant edges in biological networks. It relies on explicit eigen-decomposition and cannot be applied to large graphs.

\item ND \cite{Soheil2013network}, a blind graph denoising algorithm that solves an inverse diffusion process to remove the transitive edges. It also relies on explicit eigen-decomposition and cannot be applied to large graphs.
\item  E-Net \cite{9200731}, a non-blind graph enhancement neural network that draws inspirations from link prediction method \cite{zhang2018link}. For a fair comparison, we replace the supervision (clean graphs) used in E-Net with the noisy version. 

\item Our-ablation, an ablation of our framework that does not utilize cluster mask. It resembles VGAE \cite{kipf2016variational} with two differences: (1) we use a multi-layer perceptron in decoders while VGAE uses a non-parameterized version, and (2) we replace GCN \citep{kipf2017semi} with Heatts \citep{li2020heatts}.
\end{itemize}

\paragraph{Metrics} 

We follow image denoising \cite{pang2017graph,zeng2019deep} to use peak signal-to-noise ratio (PSNR) when the clean graph $G' = (V,A',X)$ is known.

\begin{equation}
\text{MSE} =  \frac{\sum_i^N\sum_j^i(A'_{ij} - \hat{A}_{ij})^2}{N(N-1)}.
\end{equation}

PSNR is then defined as:
\begin{equation}
\text{PSNR} = 10\log(\frac{1}{\text{MSE}}).
\end{equation}
As for the structural similarity, we leverage Weisfeiler-Lehman Graph Kernels \cite{shervashidze2011weisfeiler}.
\begin{equation}
WL = \frac{\langle \phi(\hat{A}),\phi(A') \rangle}{\sqrt{ \langle \phi(\hat{A}),\phi(\hat{A}) \rangle \langle \phi(A'),\phi(A') \rangle }},
\end{equation}
where $\langle \cdot,\cdot \rangle$ denotes dot product, $\phi(\cdot)$ represents a vector derived by Weisfeiler-Lehman Graph Kernels.

For all metrics, a greater value denotes a better performance.

\begin{table*}[t]
  \caption{Blind denoising performance comparison of different methods on benchmarks}
  \begin{center}
  \scalebox{0.7}{
  \begin{tabular}{c|cc|cc|cc|cc|cc}
    \toprule
Datasets&\multicolumn{2}{c}{IMDB-BIN}\vline& \multicolumn{2}{c}{IMDB-MULTI}\vline& \multicolumn{2}{c}{REDDIT-BIN}\vline& \multicolumn{2}{c}{MUTAG}\vline& \multicolumn{2}{c}{PTC-MR}\\
	(No. Graphs) &\multicolumn{2}{c}{1000}\vline& \multicolumn{2}{c}{1500}\vline &\multicolumn{2}{c}{2000}\vline & \multicolumn{2}{c}{188}\vline &\multicolumn{2}{c}{344}\\
	(Avg. Nodes)&\multicolumn{2}{c}{19.8}\vline&\multicolumn{2}{c}{13.0}\vline&\multicolumn{2}{c}{508.5}\vline&\multicolumn{2}{c}{17.9}\vline&\multicolumn{2}{c}{14.3}\\
	(Avg. Edges)&\multicolumn{2}{c}{193.1}\vline&\multicolumn{2}{c}{65.9}\vline&\multicolumn{2}{c}{497.8}\vline&\multicolumn{2}{c}{19.8}\vline&\multicolumn{2}{c}{14.7}\\
    \midrule
       Metrics  &PSNR&WL&PSNR&WL&PSNR&WL&PSNR&WL&PSNR&WL\\
    \midrule
    DOMINANT \cite{ding2019deep}&$50.81 \pm 0.33$&$19.79 \pm 0.41\%$&$49.29 \pm 0.65$&$54.33 \pm 0.81\%$&$74.97 \pm 1.17$&$21.88 \pm 0.69\%$&$55.09 \pm 0.47$&$25.74 \pm 1.10\%$&$42.21 \pm 0.51$&$24.88 \pm 0.45\%$\\
	ND \cite{Soheil2013network}&$47.08 \pm 0.11$&$22.47 \pm 0.14\%$&$53.22 \pm 0.05$&$55.73 \pm 0.03\%$&-&-&$43.00 \pm 0.18$&$9.37 \pm 0.22\%$&$38.12 \pm 0.27$&$22.84 \pm 0.30\%$\\
	NE \cite{wang2018network}&$51.81 \pm 0.17$&$25.87 \pm 0.09\%$&$55.40 \pm 0.21$&$57.30 \pm 0.14\%$&-&-&$54.30 \pm 0.22$&$25.65 \pm 0.18\%$&$47.92 \pm 0.05$&$22.85 \pm 0.11\%$\\
	E-Net \cite{9200731}&$49.18 \pm 0.97$&$18.38 \pm 1.15\%$&$55.57 \pm 1.01$&$56.14 \pm 1.62\%$&$87.48 \pm 1.33$&$22.54 \pm 1.07\%$&$52.42 \pm 1.58$&$30.53 \pm 1.42\%$&$50.73 \pm 1.66$&$34.51 \pm 0.92\%$\\
	\midrule
	Our-ablation&$52.75 \pm 0.48$&$21.47 \pm 0.33\%$&$55.82 \pm 0.59$&$57.85 \pm 0.55\%$&$75.35 \pm 1.10$&$23.44 \pm 0.96\%$&$55.76 \pm 0.68$&$27.90 \pm 1.12\%$&$47.52 \pm 0.44$&$33.20 \pm 0.83\%$\\
	Mask-GVAE&$\bf{ 55.74 \pm 0.51}$& $\bf{ 26.41 \pm 0.49\%}$& $\bf{ 56.86 \pm 0.66}$& $\bf{ 58.25 \pm 0.74\%}$&$\bf{ 98.41 \pm 0.71}$&$\bf{ 25.32 \pm 0.56\%}$&$\bf{ 56.72 \pm 0.31}$&$\bf{36.84 \pm 0.22\%}$
  &$\bf{ 52.06 \pm 0.56}$&$\bf{ 38.73 \pm 0.36\%}$\\
  \bottomrule
\end{tabular}
}
\end{center}
 \label{all}
\end{table*}

\begin{figure}
\begin{center}
\includegraphics [width=0.5\textwidth]{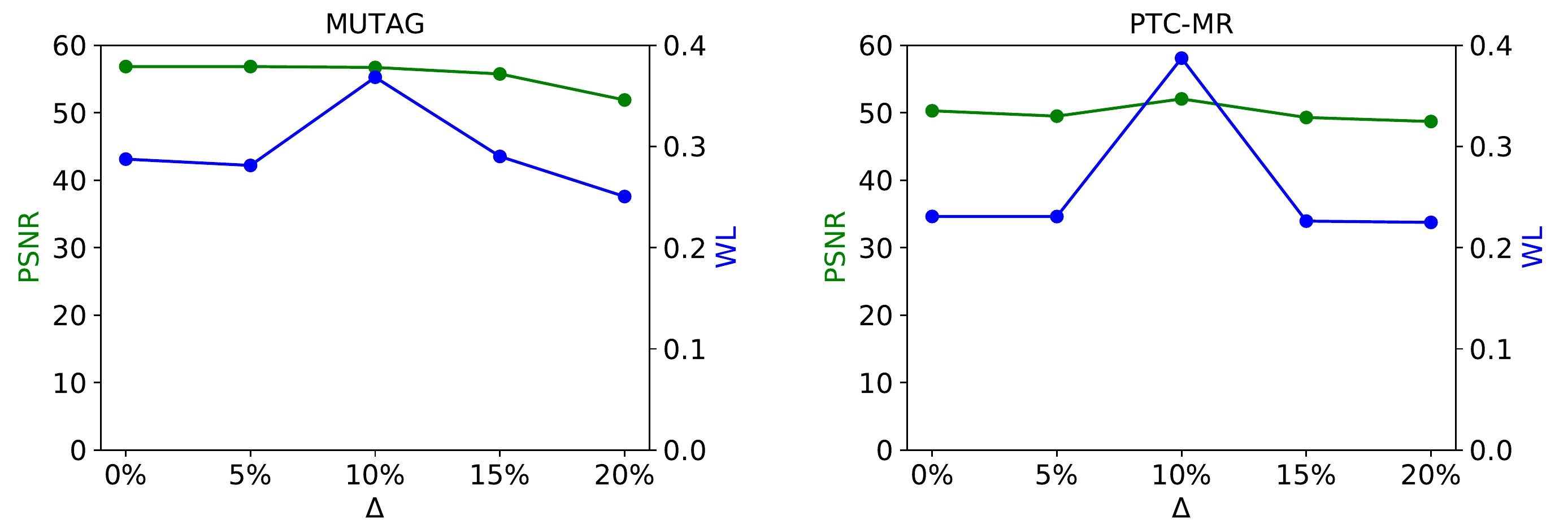}
\end{center}

\caption{Influence of the estimated noise $\Delta$ on the denoising performance of MUTAG(left) and PTC-MR(right)}
\label{budget}
\end{figure}

\paragraph{Data}
We use five graph classification benchmarks: IMDB-Binary and IMDB-Multi  \cite{Yanardag:2015} connecting actors/actresses based on movie appearance, Reddit-Binary \cite{Yanardag:2015}  connecting users through responses in Reddit online discussion, MUTAG \cite{kriege2012subgraph} containing mutagenic compound,  PTC \cite{kriege2012subgraph} containing compounds tested for carcinogenicit, as the clean graphs.  For each $A'$, we add noise to edges to produce $A$ by the following methods: (1) randomly adding 10\%  nonexistent edges, (2) randomly removing 10\% edges with respect to the existing edges in $A'$, as adopted in E-Net \cite{9200731}.   We refer to Table \ref{all} for the detailed information about the obtained noisy graphs.
\paragraph{Setup}
For Mask-GVAE, we adopt the same settings as in the experiment of graph clustering
except that we use Bayesian Information Criterion (BIC) to decide the optimal number of clusters. For all methods, we denoise the given graphs with the same budget, i.e., we add 10\% edges to recover missing edges and remove 10\% edges to delete irrelevant edges, based on the intermediate probabilistic graphs. 

\begin{table}
  \caption{Running time (in seconds) comparison of different methods}
  \label{tel}
  \begin{tabular}{cccc}
    \hline
  Data sets&IMDB-MULTI&REDDIT-BIN&PTC-MR\\
    \hline
   \textbf{DOMINANT} & 73 & 1895 & 11\\
   \textbf{ND} &41 & - & 6\\
   \textbf{NE} & 43 & - & 14\\
   \textbf{E-Net} & 216 & 2114 & 19\\
   \hline
   \textbf{Mask-GVAE} & 79 & 1043& 21\\
	  \hline
\end{tabular}
\vspace{-0.4cm}
\end{table}

\paragraph{Results}\label{xix}
Table \ref{all} lists the experimental results on the five data sets respectively. We analyse the results from the following two perspectives. 

\noindent\underline{Scalability}:  Both ND and NE suffer from a long run time and take more than 1 day on the larger data set Reddit-Binary, as they both rely on explicit eigen-decomposition. On the contrary, E-Net and our method can take advantage of GNN and avoid explicit eigen-decomposition, which makes the algorithms scalable to very large networks.

\noindent\underline{Performance}: Among all approaches, Mask-GVAE achieves the best performance on all data sets in terms of the two measures PSNR and WL similarity. It also beats Our-ablation, which shows the effectiveness of cluster mask. 

\paragraph{Running time}
Table \ref{tel} lists the running time on the three selected data sets. As can be seen, on small-sized graphs like PTC-MR, Mask-GVAE requires longer time as it needs to compute the cluster mask. On medium-sized graphs like IMDB-MULTI, the running time of Mask-GVAE is comparable to that of DOMINANT and NE/ND. On large-sized graphs like REDDIT-BIN, Mask-GVAE uses less time compared with all baselines, as the optimization of denoised graph generation in Mask-GVAE is conditioned on the observed edges.

\paragraph{Sensitivity}
We test the sensitivity of Mask-GVAE to the degree of noise and modularity \cite{newman2006modularity} in given graphs. We target random cluster graphs \cite{newman2009random} and generate 200 synthetic graphs with an average of 100 nodes. We set the degree of noise to 10\%, 20\% and 30\%. We control the modularity to be 0.05 (weak cluster structure) and 0.35 (strong cluster structure).
Table \ref{alll} lists the results. As can be seen, our solution consistently outperforms baselines in most cases, regardless of the degree of noise and cluster structure. In addition, we observe most methods perform better in PSNR on strong clustered graphs (modularity = 0.35), which shows the importance of clusters in current denoising approaches.

\paragraph{Estimating the degree of noise}
Estimating the degree of noise for the given inputs is still an open problem \cite{guo2019toward}. In this work, we use budget $\Delta$ to represent our estimation on the degree of noise in the given graphs. In this part, we evaluate how the budget $\Delta$ affects the denoising performance.  Taking MUTAG/PTC-MR with 10\% noise as an example, we vary $\Delta$ from 0\% to 20\% and plot the corresponding denoising performance in Figure \ref{budget}. As we increase $\Delta$, the curve of PSNR is quite flat, indicating our model is robust to the estimated noise on PSNR. As for WL, it first increases then drops, meaning that an appropriate noise estimation is essential for the performance of our model on structural similarity. 

\begin{table*}
  \caption{Performance comparison of different methods on the random cluster graphs}
  \scalebox{0.8}{
  \begin{tabular}{c|cccc|cccc|cccc}
    \hline
Degree of noise &\multicolumn{4}{c}{\textbf{10\%}} \vline&\multicolumn{4}{c}{\textbf{20\%}} \vline&\multicolumn{4}{c}{\textbf{30\%}}\\
Modularity&\multicolumn{2}{c}{\textbf{0.05}}&\multicolumn{2}{c}{\textbf{0.35}} \vline&\multicolumn{2}{c}{\textbf{0.05}}&\multicolumn{2}{c}{\textbf{0.35}} \vline&\multicolumn{2}{c}{\textbf{0.05}}&\multicolumn{2}{c}{\textbf{0.35}}\\
       -      &PSNR&WL&PSNR&WL&PSNR&WL&PSNR&WL&PSNR&WL&PSNR&WL\\
    \hline 
    \textbf{DOMINANT} &74.01&48.57\%&75.54&48.57\%&71.05&37.88\%&72.64&34.51\%&70.39&35.23\%&72.59&34.50\%\\
	\textbf{ND} &72.70&48.19\%&75.67&46.60\%&71.30&27.20\%&73.74&28.73\%&70.37&18.01\%&71.98&17.20\%\\
	\textbf{NE} &74.48&46.67\%&76.82&44.24\%&71.98&35.06\%&73.66&33.66\%&70.33&27.67\%&72.10&28.07\%\\
	\textbf{E-Net} &75.91&48.17\%&77.58&47.88\%&72.24&38.94\%&74.09&36.47\%&70.36&34.99\%&72.51&33.09\%\\
	\hline
	\textbf{Mask-GVAE} &\textbf{76.32}&\textbf{48.94\%}&\textbf{78.01}&\textbf{50.07\%}&\textbf{72.73}&\textbf{48.37\%}&\textbf{74.66}&\textbf{36.82\%}&\textbf{70.49}&\textbf{36.14\%}&\textbf{72.68}&\textbf{35.99\%}\\
  \hline
\end{tabular}
}
 \label{alll}
\vspace{-0.4cm}
\end{table*}

\begin{figure}
\centering
\includegraphics [width=0.48\textwidth]{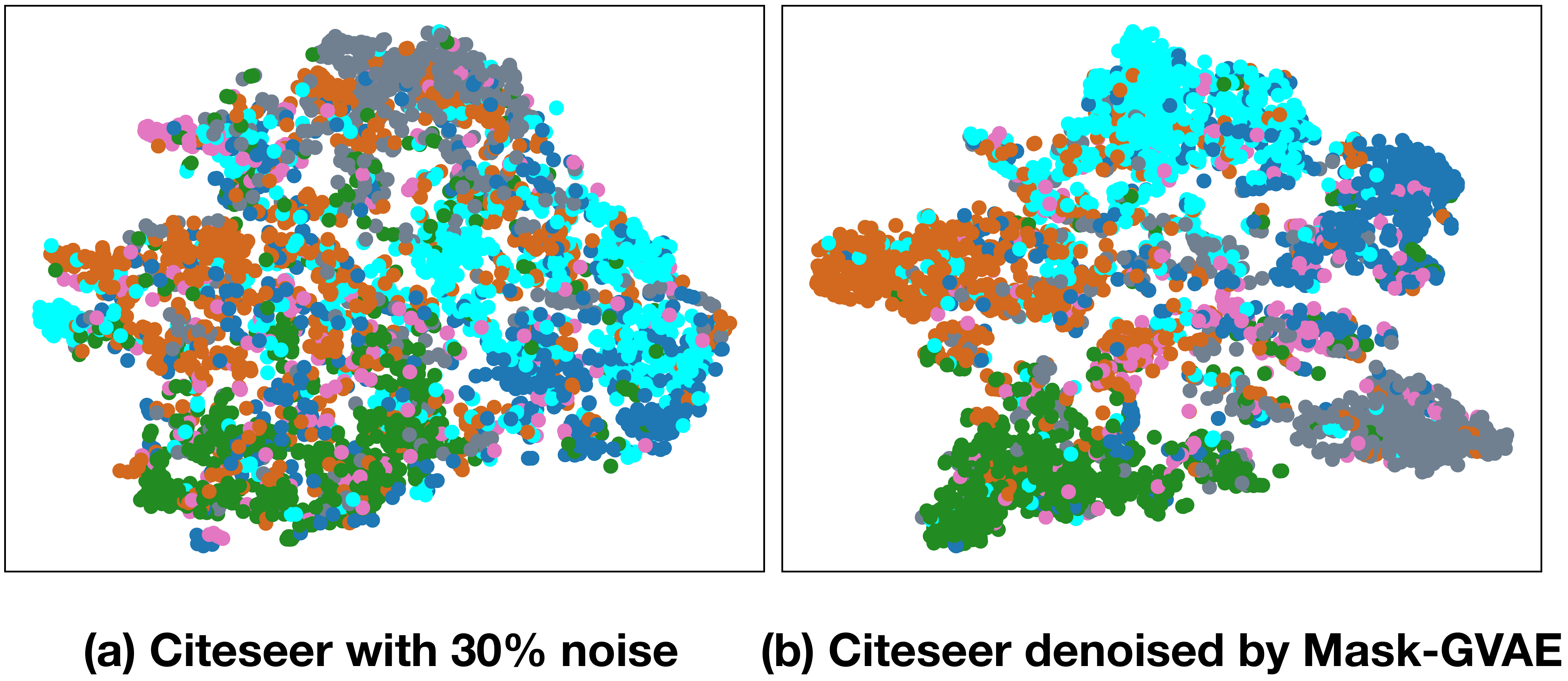}
\caption{ Two-dimensional visualization of node embeddings before and after the denoising of Mask-GVAE on Citeseer. }
\label{fig.vi}
\end{figure}

\paragraph{Visualization}
We target Citeseer and add 30\% noise. In Figure \ref{fig.vi}, we derive node embeddings before/after Mask-GVAE by Node2vec \cite{grover2016node2vec} and project the embeddings into a two-dimensional space with t-SNE, in which different colors denote different classes.  For noised Citeseer, nodes of different classes are mixed up, as reflected by the geometric distances between different colors. Mask-GVAE can make a distinct difference between classes.   

\begin{figure}
\begin{center}
\includegraphics [width=0.5\textwidth,scale=1]{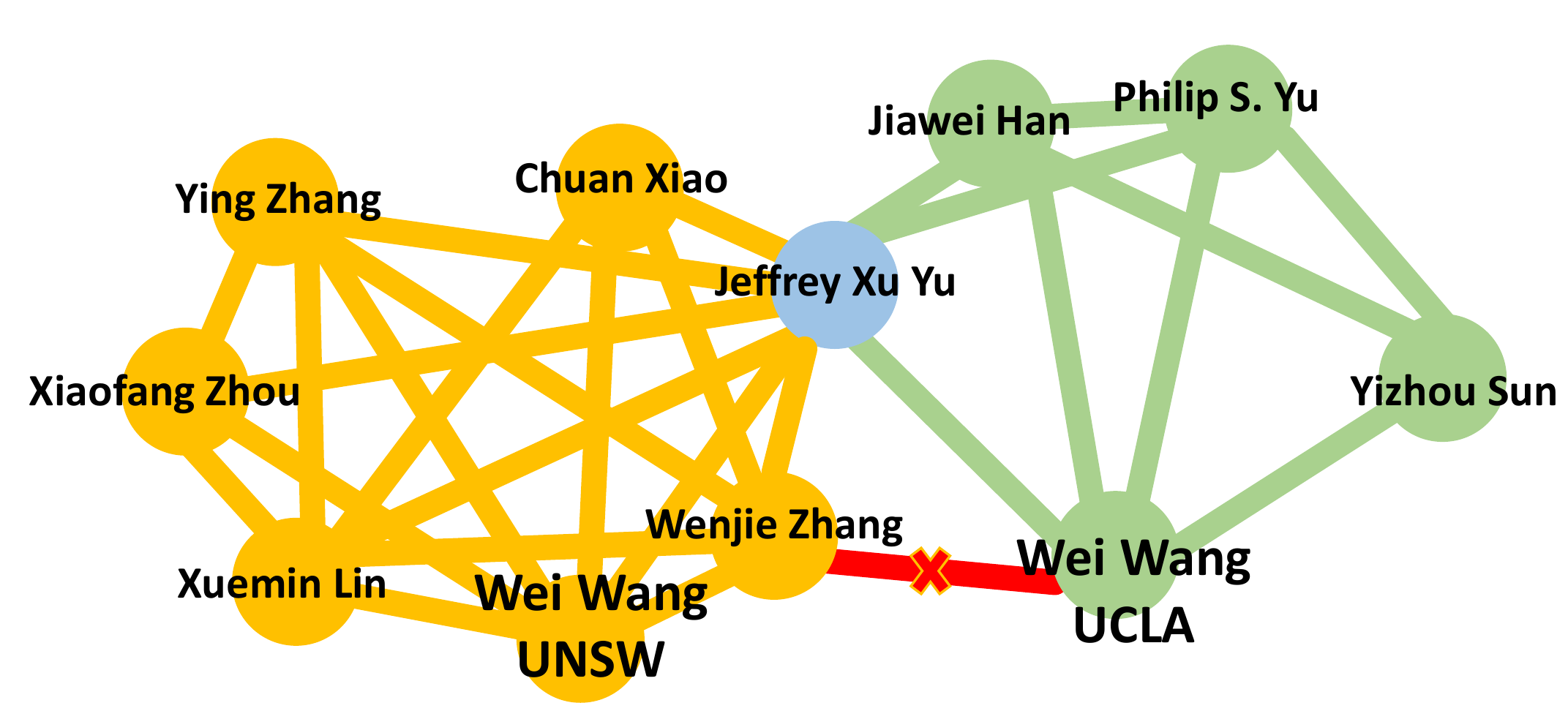}
\end{center}

\caption{A case to show how Mask-GVAE denoises graphs}
\label{fig.va}
\end{figure}

\paragraph{Case study}
To have a better understanding of how Mask-GVAE works, we target a subgraph of co-authorship network  which consists of 11 scholars in the area of Data Base (DB) and Data Mining (DM). The subgraph is based on Google Scholar\footnote{https://scholar.google.com/} and constructed as follows: (1) for each scholar, we use a one-hot representation with 300 dimensions encoding his/her research keywords, and (2) we construct an adjacency matrix $A$ by denoting $A_{ij} = 1$ and $A_{ji} = 1$ if the two scholars have co-authored papers. 
We apply Mask-GVAE on this subgraph and set $K=2$ and $\Delta = 1$ to remove irrelevant edges.  As can be seen from Figure \ref{fig.va}, Mask-GVAE identifies the connection between ``Wenjie Zhang'' and ``Wei Wang (UCLA)'' as the noisy edge.  We further notice that the noisy edge is constructed based on paper \cite{li2019approximate}, which is actually co-authored by ``Wenjie Zhang'' and ``Wei Wang (UNSW)''.  One
possible explanation is that Google Scholar fails to distinguish  ``Wei Wang (UNSW)'' and ``Wei Wang (UCLA)'', as a result the paper \cite{li2019approximate} appears on the Google Scholar page of ``Wei Wang (UCLA)'' mistakenly.

As a comparison, when we apply ND, NE and E-Net on this subgraph, only NE can pick up the noisy connection, while ND and E-Net cannot. One
possibility is Mask-GVAE and NE are designed to keep the cluster structure. But ND targets ``Jeffery Xu Yu” and ``Wei Wang (UCLA)” and E-Net targets ``Jeffery Xu Yu” and ``Jiawei Han”, both of which are true connections.

\section{Related work}\label{sec.related}
\noindent\textbf{Graph Laplacian in denoising tasks}.  Several studies have utilized graph Laplacian in various denoising tasks, including image denoising \cite{pang2017graph,zeng2019deep}, signal denoising \cite{dong2016learning}, and 3D point cloud denoising \cite{hu2019feature}.  As graph structures are not readily available in these domains, these studies differ in how they construct graph structures, i.e., structures are only created as an auxiliary source to recover a high quality image/signal/3D point cloud.  Only recently, E-Net \cite{9200731} has proposed to adopt Graph Laplacian regularization in non-blind graph denoising tasks, in order to restore a denoised graph with \emph{global smoothing} and \emph{sparsity}.

\noindent\textbf{Graph denoising}.  Although there have been many studies on image denoising, graph denoising has been studied less, in particular, the study of blind denoising large discrete graphs is still lacking. It is worth noting the study of signal denoising on graphs \cite{chen2014signal,padilla2017dfs} is different from the study of graph structure denoising. When it comes to structure denoising, ND \cite{Soheil2013network} formulates the problem as the inverse of network convolution, and introduces an algorithm that removes the combined effect of all indirect paths by exploiting eigen-decomposition. NE \cite{wang2018network} recovers the graph structure based on diffusion algorithms and follows the intuition that nodes connected by high-weight paths are more likely to have direct edges. Low-rank estimation \cite{hsieh2012low} and sparse matrix estimation \cite{richard2012estimation} assume the given graph is incomplete and noisy, thus they aim to recover the structure with the property of low-rank/sparsity. \cite{morris2004denoising} infers the hidden graph structures based on the heuristic that a set of hidden, constituent subgraphs are combined to generate the observed graph. Inspired by link prediction \cite{zhang2018link}, E-Net \cite{9200731} enhances the quality of graph structure via exploiting subgraph characteristics and GNNs. Moreover, E-NET requires supervision of the clean graphs, which is different from our blind setting.

\noindent\textbf{Utilization of substructure}. It is a common practice to utilize certain substructures (i.e., clusters or
subgraphs) in denoising tasks \cite{morris2004denoising,9200731,Soheil2013network} or other related areas \cite{zhang2018link, chiang2019cluster}. The underlying ideas can be generally classified into two groups. The first group is related to scalability, i.e.,  substructures make the algorithms scalable to very large networks. Representative works include ND \cite{Soheil2013network}, E-Net \cite{9200731}, Cluster-GCN \cite{chiang2019cluster}. The second group considers substructures can provide meaningful context for the tasks at hand. For example, SEAL \cite{zhang2018link} shows that local subgraphs reserve
rich information related to link existence. However, all these methods are heuristic methods and shed little light on when these substructure based ideas fail. In this work, we prove that in order to make substructure heuristics work, the given graph should have a distinct cluster structure. Moreover, it is the robustness of low eigenvectors in graph Laplacian matrix that lays the foundation of these heuristics.

\noindent\textbf{Discrete graph generation}.
Directly generating discrete graph with gradient descent methods is intractable. In this regard, several efforts have been made to bypass the difficulty. \cite{bello2016neural} learns a policy network with reinforcement learning, \cite{jang2016categorical} approximates the discrete data by Gumbel distribution, \cite{simonovsky2018graphvae} circumvents the problem by formulating the loss on a probabilistic graph and drawing discrete graphs thereafter, which we follow in this work.

\section{Conclusion}\label{sec.con}
In this paper, we present Mask-GVAE, the first variational generative model for blind denoising large discrete graphs.  Given the huge search space of selecting proper candidates, we decompose the graph into several subgraphs and generate smoothed clusters in a variational manner, which is based on the assumption that low eigenvectors are robust against random noise. The effectiveness of Mask-GVAE is validated on five graph benchmarks, with a significant improvement on PSNR and WL similarity. 

\begin{acks}
The work described in this paper was supported by grants from the Research Grants Council of the Hong Kong Special Administrative Region, China [Project No.: CUHK 14205617] and [Project No.: CUHK 14205618], Huawei Technologies Research and Development Fund, and NSFC Grant No. U1936205.
\end{acks}

%% file: appendices.tex
\section{Proof of  Proposition 4.1}\label{a.d}

\begin{lemma}
\label{lm:lam3}
If Assumption \ref{thm:ass} holds, then $P(\lambda_3 \geq (1 - \frac{1}{\epsilon})\lambda'_3) \geq 1 - N^{-\frac{1}{8}}$.
\end{lemma}

We first prove lemma \ref{lm:lam3}. Let $E$ be the set of edges of $G'$,  $F$ be the set of edges removed from $E$, $E_e = E_{ii} + E_{jj} - E_{ij} - E_{ji}$, $E_{ij}$ is the matrix with 1 for position $(i,j)$  and 0 for others, and $E_F = \sum_{e \in F} E_e$, if $(i,j) \in E$. 
Let $X_e$ be an indicator random variable of the event that $e \in F$. 
Thus, $E_F = \sum_{e \in E} = X_e E_e$, $E_F$ is the Laplacian matrix of the induced graph  $F$.
It is easy to get $P[X_e = 1] = q$, $\mathbb{E}[E_F] = q L'$. 
Let $\mu = \lambda_{max}(\mathbb{E}[E_F]) = q\lambda'_N$.
Furthermore, note that $0 \leq \lambda(X_e E_e) \leq \lambda_{max}(X_e E_e) \leq 2$. 
By the matrix Chernoff bound in Peng and Yoshida \cite{peng2002aver}, for any $s > 0$, we have
\begin{equation*}
P(\lambda_{max}(E_F) \geq (1 + s)\mu) \leq N \Big( \frac{\exp(s)}{(1 + s)^{1 + s}} \Big)^{\frac{1}{2}\mu}. 
\end{equation*}
Set $1 + s = \frac{25}{2}, (1 + s)\mu = \frac{3}{2}\log N$, then $P(\lambda_{max}(E_F) \geq (1 + s)\mu) \leq N^{-\frac{1}{8}}$. 
Thus with probability at least $1 - N^{-\frac{1}{8}}$, $\lambda_{max}(E_F) \leq \max(\frac{25}{2}\mu, \frac{3}{2} \log N) = \max(\frac{25}{2}q\lambda'_N, \frac{3}{2} \log N) \leq \frac{\lambda'_3}{\epsilon}$ holds.
Due to the fact that $\lambda_3 \geq \lambda'_3 - \lambda_{max}(E_F)$, $P(\lambda_3 \geq (1 - \frac{1}{\epsilon})\lambda'_3) \geq 1 - N^{-\frac{1}{8}}$ holds.

	With lemma \ref{lm:lam3}, we prove Proposition 4.1  based on the well-known Davis-Kahan perturbation theorem.
	
	It is easy to show that	$\mathbb{E}[\Vert L - L' \Vert_F] = q \Vert L' \Vert_F$.
	
	Define $\delta = \min\{|\lambda_j - \lambda'_2|, j\neq 2\}$, we can get the following bound according to Davis-Kahan theorem,  
	
	\begin{equation*}
		\mathbb{E}[\sin(\angle(\mathbf{u}_2, \mathbf{u'}_2))] \leq  \mathbb{E}[\frac{1}{\delta}\Vert L - L' \Vert_F] = \frac{1}{\delta}(q \Vert L' \Vert_F). 
	\end{equation*}
	
	Easy to check $\Vert L' \Vert_F^2 = O(\max(N^{\chi}, N^{\varphi}))$.
	
	By Lemma \ref{lm:lam3},  we have $P(\lambda_3 \geq (1 - \frac{1}{\epsilon})\lambda'_3) \geq 1 - N^{-\frac{1}{8}}$. 
	Note that if $\frac{2\epsilon}{\epsilon-1}\lambda'_2 < \lambda'_3$, then $\delta = \lambda'_2$,  thus 
	
	if $\beta > \frac{1}{2} \max(\chi, \varphi)$, then $\mathbb{E} [\sin(\angle(\mathbf{u}_2, \mathbf{u'}_2))] \rightarrow 0$.
	
	if $\beta = \frac{1}{2} \max(\chi, \varphi)$, then $\mathbb{E}[\sin(\angle(\mathbf{u}_2, \mathbf{u'}_2))] \leq \frac{1}{\kappa}$.

%% file: bibil_1.bbl

%% file: sample-sigconf.bbl
\begin{thebibliography}{68}


\ifx \showCODEN    \undefined \def \showCODEN     #1{\unskip}     \fi
\ifx \showDOI      \undefined \def \showDOI       #1{#1}\fi
\ifx \showISBNx    \undefined \def \showISBNx     #1{\unskip}     \fi
\ifx \showISBNxiii \undefined \def \showISBNxiii  #1{\unskip}     \fi
\ifx \showISSN     \undefined \def \showISSN      #1{\unskip}     \fi
\ifx \showLCCN     \undefined \def \showLCCN      #1{\unskip}     \fi
\ifx \shownote     \undefined \def \shownote      #1{#1}          \fi
\ifx \showarticletitle \undefined \def \showarticletitle #1{#1}   \fi
\ifx \showURL      \undefined \def \showURL       {\relax}        \fi
\providecommand\bibfield[2]{#2}
\providecommand\bibinfo[2]{#2}
\providecommand\natexlab[1]{#1}
\providecommand\showeprint[2][]{arXiv:#2}

\bibitem[\protect\citeauthoryear{Ahn, Bagrow, and Lehmann}{Ahn
  et~al\mbox{.}}{2010}]%
        {ahn2010link}
\bibfield{author}{\bibinfo{person}{Yong-Yeol Ahn}, \bibinfo{person}{James~P
  Bagrow}, {and} \bibinfo{person}{Sune Lehmann}.}
  \bibinfo{year}{2010}\natexlab{}.
\newblock \showarticletitle{Link communities reveal multiscale complexity in
  networks}.
\newblock \bibinfo{journal}{\emph{Nature}} \bibinfo{volume}{466},
  \bibinfo{number}{7307} (\bibinfo{year}{2010}), \bibinfo{pages}{761--764}.
\newblock


\bibitem[\protect\citeauthoryear{Bello, Pham, Le, Norouzi, and Bengio}{Bello
  et~al\mbox{.}}{2017}]%
        {bello2016neural}
\bibfield{author}{\bibinfo{person}{Irwan Bello}, \bibinfo{person}{Hieu Pham},
  \bibinfo{person}{Quoc~V Le}, \bibinfo{person}{Mohammad Norouzi}, {and}
  \bibinfo{person}{Samy Bengio}.} \bibinfo{year}{2017}\natexlab{}.
\newblock \showarticletitle{Neural combinatorial optimization with
  reinforcement learning}.
\newblock \bibinfo{journal}{\emph{International Conference on Learning
  Representations (ICLR)}} (\bibinfo{year}{2017}).
\newblock


\bibitem[\protect\citeauthoryear{Cavallari, Zheng, Cai, Chang, and
  Cambria}{Cavallari et~al\mbox{.}}{2017}]%
        {cavallari2017learning}
\bibfield{author}{\bibinfo{person}{Sandro Cavallari},
  \bibinfo{person}{Vincent~W Zheng}, \bibinfo{person}{Hongyun Cai},
  \bibinfo{person}{Kevin Chen-Chuan Chang}, {and} \bibinfo{person}{Erik
  Cambria}.} \bibinfo{year}{2017}\natexlab{}.
\newblock \showarticletitle{Learning community embedding with community
  detection and node embedding on graphs}. In
  \bibinfo{booktitle}{\emph{Proceedings of the 2017 ACM on Conference on
  Information and Knowledge Management (CIKM)}}. ACM,
  \bibinfo{pages}{377--386}.
\newblock


\bibitem[\protect\citeauthoryear{Chen, Sandryhaila, Moura, and Kovacevic}{Chen
  et~al\mbox{.}}{2014}]%
        {chen2014signal}
\bibfield{author}{\bibinfo{person}{Siheng Chen}, \bibinfo{person}{Aliaksei
  Sandryhaila}, \bibinfo{person}{Jos{\'e}~MF Moura}, {and}
  \bibinfo{person}{Jelena Kovacevic}.} \bibinfo{year}{2014}\natexlab{}.
\newblock \showarticletitle{Signal denoising on graphs via graph filtering}. In
  \bibinfo{booktitle}{\emph{2014 IEEE Global Conference on Signal and
  Information Processing (GlobalSIP)}}. IEEE, \bibinfo{pages}{872--876}.
\newblock


\bibitem[\protect\citeauthoryear{Chen, Li, and Bruna}{Chen
  et~al\mbox{.}}{2019}]%
        {chen2017supervised}
\bibfield{author}{\bibinfo{person}{Zhengdao Chen}, \bibinfo{person}{Xiang Li},
  {and} \bibinfo{person}{Joan Bruna}.} \bibinfo{year}{2019}\natexlab{}.
\newblock \showarticletitle{Supervised community detection with line graph
  neural networks}.
\newblock \bibinfo{journal}{\emph{International Conference on Learning
  Representations (ICLR)}} (\bibinfo{year}{2019}).
\newblock


\bibitem[\protect\citeauthoryear{Chiang, Liu, Si, Li, Bengio, and Hsieh}{Chiang
  et~al\mbox{.}}{2019}]%
        {chiang2019cluster}
\bibfield{author}{\bibinfo{person}{Wei-Lin Chiang}, \bibinfo{person}{Xuanqing
  Liu}, \bibinfo{person}{Si Si}, \bibinfo{person}{Yang Li},
  \bibinfo{person}{Samy Bengio}, {and} \bibinfo{person}{Cho-Jui Hsieh}.}
  \bibinfo{year}{2019}\natexlab{}.
\newblock \showarticletitle{Cluster-GCN: An efficient algorithm for training
  deep and large graph convolutional networks}. In
  \bibinfo{booktitle}{\emph{Proceedings of the 25th ACM SIGKDD International
  Conference on Knowledge Discovery \& Data Mining}}.
  \bibinfo{pages}{257--266}.
\newblock


\bibitem[\protect\citeauthoryear{Chorowski, Weiss, Bengio, and van~den
  Oord}{Chorowski et~al\mbox{.}}{2019}]%
        {chorowski2019unsupervised}
\bibfield{author}{\bibinfo{person}{Jan Chorowski}, \bibinfo{person}{Ron~J
  Weiss}, \bibinfo{person}{Samy Bengio}, {and} \bibinfo{person}{A{\"a}ron
  van~den Oord}.} \bibinfo{year}{2019}\natexlab{}.
\newblock \showarticletitle{Unsupervised speech representation learning using
  wavenet autoencoders}.
\newblock \bibinfo{journal}{\emph{IEEE/ACM transactions on audio, speech, and
  language processing}} \bibinfo{volume}{27}, \bibinfo{number}{12}
  (\bibinfo{year}{2019}), \bibinfo{pages}{2041--2053}.
\newblock


\bibitem[\protect\citeauthoryear{Cohen, Rosenfeld, and Kolter}{Cohen
  et~al\mbox{.}}{2019}]%
        {cohen2019certified}
\bibfield{author}{\bibinfo{person}{Jeremy~M Cohen}, \bibinfo{person}{Elan
  Rosenfeld}, {and} \bibinfo{person}{J~Zico Kolter}.}
  \bibinfo{year}{2019}\natexlab{}.
\newblock \showarticletitle{Certified adversarial robustness via randomized
  smoothing}.
\newblock \bibinfo{journal}{\emph{ICML}} (\bibinfo{year}{2019}).
\newblock


\bibitem[\protect\citeauthoryear{Ding, Li, Bhanushali, and Liu}{Ding
  et~al\mbox{.}}{2019}]%
        {ding2019deep}
\bibfield{author}{\bibinfo{person}{Kaize Ding}, \bibinfo{person}{Jundong Li},
  \bibinfo{person}{Rohit Bhanushali}, {and} \bibinfo{person}{Huan Liu}.}
  \bibinfo{year}{2019}\natexlab{}.
\newblock \showarticletitle{Deep anomaly detection on attributed networks}. In
  \bibinfo{booktitle}{\emph{Proceedings of the 2019 SIAM International
  Conference on Data Mining}}. SIAM, \bibinfo{pages}{594--602}.
\newblock


\bibitem[\protect\citeauthoryear{Dobson and Doig}{Dobson and Doig}{2003}]%
        {dobson2003distinguishing}
\bibfield{author}{\bibinfo{person}{P.~D. Dobson} {and} \bibinfo{person}{A.~J.
  Doig}.} \bibinfo{year}{2003}\natexlab{}.
\newblock \showarticletitle{Distinguishing enzyme structures from non-enzymes
  without alignments}.
\newblock \bibinfo{journal}{\emph{Journal of molecular biology}}
  \bibinfo{volume}{330}, \bibinfo{number}{4} (\bibinfo{year}{2003}),
  \bibinfo{pages}{771--783}.
\newblock


\bibitem[\protect\citeauthoryear{Dong, Thanou, Frossard, and
  Vandergheynst}{Dong et~al\mbox{.}}{2016}]%
        {dong2016learning}
\bibfield{author}{\bibinfo{person}{Xiaowen Dong}, \bibinfo{person}{Dorina
  Thanou}, \bibinfo{person}{Pascal Frossard}, {and} \bibinfo{person}{Pierre
  Vandergheynst}.} \bibinfo{year}{2016}\natexlab{}.
\newblock \showarticletitle{Learning Laplacian matrix in smooth graph signal
  representations}.
\newblock \bibinfo{journal}{\emph{IEEE Transactions on Signal Processing}}
  \bibinfo{volume}{64}, \bibinfo{number}{23} (\bibinfo{year}{2016}),
  \bibinfo{pages}{6160--6173}.
\newblock


\bibitem[\protect\citeauthoryear{Eldridge, Belkin, and Wang}{Eldridge
  et~al\mbox{.}}{2018}]%
        {eldridge2017unperturbed}
\bibfield{author}{\bibinfo{person}{Justin Eldridge}, \bibinfo{person}{Mikhail
  Belkin}, {and} \bibinfo{person}{Yusu Wang}.} \bibinfo{year}{2018}\natexlab{}.
\newblock \showarticletitle{Unperturbed: spectral analysis beyond Davis-Kahan}.
  In \bibinfo{booktitle}{\emph{Proceedings of Algorithmic Learning Theory}},
  Vol.~\bibinfo{volume}{83}. \bibinfo{publisher}{PMLR},
  \bibinfo{pages}{321--358}.
\newblock


\bibitem[\protect\citeauthoryear{Feizi, Marbach, Médard, and Kellis}{Feizi
  et~al\mbox{.}}{2013}]%
        {Soheil2013network}
\bibfield{author}{\bibinfo{person}{Soheil Feizi}, \bibinfo{person}{Daniel
  Marbach}, \bibinfo{person}{Muriel Médard}, {and} \bibinfo{person}{Manolis
  Kellis}.} \bibinfo{year}{2013}\natexlab{}.
\newblock \showarticletitle{Network deconvolution as a general method to
  distinguish direct dependencies in networks}.
\newblock \bibinfo{journal}{\emph{Nature Biotechnology}}
  (\bibinfo{year}{2013}), \bibinfo{pages}{726–733}.
\newblock


\bibitem[\protect\citeauthoryear{Felzenszwalb and Huttenlocher}{Felzenszwalb
  and Huttenlocher}{2004}]%
        {felzenszwalb2004efficient}
\bibfield{author}{\bibinfo{person}{Pedro~F Felzenszwalb} {and}
  \bibinfo{person}{Daniel~P Huttenlocher}.} \bibinfo{year}{2004}\natexlab{}.
\newblock \showarticletitle{Efficient graph-based image segmentation}.
\newblock \bibinfo{journal}{\emph{International journal of computer vision}}
  \bibinfo{volume}{59}, \bibinfo{number}{2} (\bibinfo{year}{2004}),
  \bibinfo{pages}{167--181}.
\newblock


\bibitem[\protect\citeauthoryear{Grover and Leskovec}{Grover and
  Leskovec}{2016}]%
        {grover2016node2vec}
\bibfield{author}{\bibinfo{person}{Aditya Grover} {and} \bibinfo{person}{Jure
  Leskovec}.} \bibinfo{year}{2016}\natexlab{}.
\newblock \showarticletitle{node2vec: Scalable feature learning for networks}.
  In \bibinfo{booktitle}{\emph{Proceedings of the 22nd ACM SIGKDD International
  Conference on Knowledge Discovery and Data Mining (SIGKDD)}}.
  \bibinfo{pages}{855--864}.
\newblock


\bibitem[\protect\citeauthoryear{Guo, Yan, Zhang, Zuo, and Zhang}{Guo
  et~al\mbox{.}}{2019}]%
        {guo2019toward}
\bibfield{author}{\bibinfo{person}{Shi Guo}, \bibinfo{person}{Zifei Yan},
  \bibinfo{person}{Kai Zhang}, \bibinfo{person}{Wangmeng Zuo}, {and}
  \bibinfo{person}{Lei Zhang}.} \bibinfo{year}{2019}\natexlab{}.
\newblock \showarticletitle{Toward convolutional blind denoising of real
  photographs}. In \bibinfo{booktitle}{\emph{Proceedings of the IEEE Conference
  on Computer Vision and Pattern Recognition}}. \bibinfo{pages}{1712--1722}.
\newblock


\bibitem[\protect\citeauthoryear{Hsieh, Chiang, and Dhillon}{Hsieh
  et~al\mbox{.}}{2012}]%
        {hsieh2012low}
\bibfield{author}{\bibinfo{person}{Cho-Jui Hsieh}, \bibinfo{person}{Kai-Yang
  Chiang}, {and} \bibinfo{person}{Inderjit~S Dhillon}.}
  \bibinfo{year}{2012}\natexlab{}.
\newblock \showarticletitle{Low rank modeling of signed networks}. In
  \bibinfo{booktitle}{\emph{Proceedings of the 18th ACM SIGKDD International
  Conference on Knowledge Discovery and Data Mining (SIGKDD)}}.
  \bibinfo{pages}{507--515}.
\newblock


\bibitem[\protect\citeauthoryear{Hu, Gao, Cheung, and Guo}{Hu
  et~al\mbox{.}}{2020}]%
        {hu2019feature}
\bibfield{author}{\bibinfo{person}{Wei Hu}, \bibinfo{person}{Xiang Gao},
  \bibinfo{person}{Gene Cheung}, {and} \bibinfo{person}{Zongming Guo}.}
  \bibinfo{year}{2020}\natexlab{}.
\newblock \showarticletitle{Feature graph learning for 3d point cloud
  denoising}.
\newblock \bibinfo{journal}{\emph{IEEE Transactions on Signal Processing}}
  \bibinfo{volume}{68} (\bibinfo{year}{2020}), \bibinfo{pages}{2841--2856}.
\newblock


\bibitem[\protect\citeauthoryear{Huang, Yan, Taft, and Jordan}{Huang
  et~al\mbox{.}}{2009}]%
        {huang2009spectral}
\bibfield{author}{\bibinfo{person}{Ling Huang}, \bibinfo{person}{Donghui Yan},
  \bibinfo{person}{Nina Taft}, {and} \bibinfo{person}{Michael~I Jordan}.}
  \bibinfo{year}{2009}\natexlab{}.
\newblock \showarticletitle{Spectral clustering with perturbed data}. In
  \bibinfo{booktitle}{\emph{Proceedings of the 21st International Conference on
  Neural Information Processing Systems (NeurIPS)}}. \bibinfo{pages}{705--712}.
\newblock


\bibitem[\protect\citeauthoryear{Im, Ahn, Memisevic, and Bengio}{Im
  et~al\mbox{.}}{2017}]%
        {im2017denoising}
\bibfield{author}{\bibinfo{person}{Daniel Im~Jiwoong Im},
  \bibinfo{person}{Sungjin Ahn}, \bibinfo{person}{Roland Memisevic}, {and}
  \bibinfo{person}{Yoshua Bengio}.} \bibinfo{year}{2017}\natexlab{}.
\newblock \showarticletitle{Denoising criterion for variational auto-encoding
  framework}. In \bibinfo{booktitle}{\emph{Thirty-First AAAI Conference on
  Artificial Intelligence}}. \bibinfo{pages}{2059--2065}.
\newblock


\bibitem[\protect\citeauthoryear{Jang, Gu, and Poole}{Jang
  et~al\mbox{.}}{2016}]%
        {jang2016categorical}
\bibfield{author}{\bibinfo{person}{Eric Jang}, \bibinfo{person}{Shixiang Gu},
  {and} \bibinfo{person}{Ben Poole}.} \bibinfo{year}{2016}\natexlab{}.
\newblock \showarticletitle{Categorical reparameterization with
  gumbel-softmax}.
\newblock \bibinfo{journal}{\emph{International Conference on Learning
  Representations (ICLR)}} (\bibinfo{year}{2016}).
\newblock


\bibitem[\protect\citeauthoryear{Jia, Wang, Cao, and Gong}{Jia
  et~al\mbox{.}}{2020}]%
        {jia2020certified}
\bibfield{author}{\bibinfo{person}{Jinyuan Jia}, \bibinfo{person}{Binghui
  Wang}, \bibinfo{person}{Xiaoyu Cao}, {and} \bibinfo{person}{Neil~Zhenqiang
  Gong}.} \bibinfo{year}{2020}\natexlab{}.
\newblock \showarticletitle{Certified Robustness of Community Detection against
  Adversarial Structural Perturbation via Randomized Smoothing}. In
  \bibinfo{booktitle}{\emph{Proceedings of The Web Conference 2020}}.
  \bibinfo{pages}{2718--2724}.
\newblock


\bibitem[\protect\citeauthoryear{Karrer, Levina, and Newman}{Karrer
  et~al\mbox{.}}{2008}]%
        {karrer2008robustness}
\bibfield{author}{\bibinfo{person}{Brian Karrer}, \bibinfo{person}{Elizaveta
  Levina}, {and} \bibinfo{person}{Mark~EJ Newman}.}
  \bibinfo{year}{2008}\natexlab{}.
\newblock \showarticletitle{Robustness of community structure in networks}.
\newblock \bibinfo{journal}{\emph{Physical review E}} \bibinfo{volume}{77},
  \bibinfo{number}{4} (\bibinfo{year}{2008}), \bibinfo{pages}{046119}.
\newblock


\bibitem[\protect\citeauthoryear{Kim and Diesner}{Kim and Diesner}{2016}]%
        {kim2016distortive}
\bibfield{author}{\bibinfo{person}{Jinseok Kim} {and} \bibinfo{person}{Jana
  Diesner}.} \bibinfo{year}{2016}\natexlab{}.
\newblock \showarticletitle{Distortive effects of initial-based name
  disambiguation on measurements of large-scale coauthorship networks}.
\newblock \bibinfo{journal}{\emph{Journal of the Association for Information
  Science and Technology}} \bibinfo{volume}{67}, \bibinfo{number}{6}
  (\bibinfo{year}{2016}), \bibinfo{pages}{1446--1461}.
\newblock


\bibitem[\protect\citeauthoryear{Kingma, Salimans, Jozefowicz, Chen, Sutskever,
  and Welling}{Kingma et~al\mbox{.}}{2016}]%
        {kingma2016improved}
\bibfield{author}{\bibinfo{person}{Durk~P Kingma}, \bibinfo{person}{Tim
  Salimans}, \bibinfo{person}{Rafal Jozefowicz}, \bibinfo{person}{Xi Chen},
  \bibinfo{person}{Ilya Sutskever}, {and} \bibinfo{person}{Max Welling}.}
  \bibinfo{year}{2016}\natexlab{}.
\newblock \showarticletitle{Improved variational inference with inverse
  autoregressive flow}. In \bibinfo{booktitle}{\emph{Advances in neural
  information processing systems}}. \bibinfo{pages}{4743--4751}.
\newblock


\bibitem[\protect\citeauthoryear{Kingma and Welling}{Kingma and
  Welling}{2013}]%
        {kingma2013auto}
\bibfield{author}{\bibinfo{person}{Diederik~P Kingma} {and}
  \bibinfo{person}{Max Welling}.} \bibinfo{year}{2013}\natexlab{}.
\newblock \showarticletitle{Auto-encoding variational bayes}.
\newblock \bibinfo{journal}{\emph{International Conference on Learning
  Representations (ICLR)}} (\bibinfo{year}{2013}).
\newblock


\bibitem[\protect\citeauthoryear{Kipf and Welling}{Kipf and Welling}{2016}]%
        {kipf2016variational}
\bibfield{author}{\bibinfo{person}{Thomas~N Kipf} {and} \bibinfo{person}{Max
  Welling}.} \bibinfo{year}{2016}\natexlab{}.
\newblock \showarticletitle{Variational Graph Auto-Encoders}.
\newblock \bibinfo{journal}{\emph{NeurIPS Workshop on Bayesian Deep Learning}}
  (\bibinfo{year}{2016}).
\newblock


\bibitem[\protect\citeauthoryear{Kipf and Welling}{Kipf and Welling}{2017}]%
        {kipf2017semi}
\bibfield{author}{\bibinfo{person}{Thomas~N. Kipf} {and} \bibinfo{person}{Max
  Welling}.} \bibinfo{year}{2017}\natexlab{}.
\newblock \showarticletitle{Semi-Supervised Classification with Graph
  Convolutional Networks}. In \bibinfo{booktitle}{\emph{International
  Conference on Learning Representations (ICLR)}}.
\newblock


\bibitem[\protect\citeauthoryear{Kriege and Mutzel}{Kriege and Mutzel}{2012}]%
        {kriege2012subgraph}
\bibfield{author}{\bibinfo{person}{Nils Kriege} {and} \bibinfo{person}{Petra
  Mutzel}.} \bibinfo{year}{2012}\natexlab{}.
\newblock \showarticletitle{Subgraph matching kernels for attributed graphs}.
\newblock \bibinfo{journal}{\emph{ICML}} (\bibinfo{year}{2012}),
  \bibinfo{pages}{291–298}.
\newblock


\bibitem[\protect\citeauthoryear{Kusner, Paige, and
  Hern{\'a}ndez-Lobato}{Kusner et~al\mbox{.}}{2017}]%
        {kusner2017grammar}
\bibfield{author}{\bibinfo{person}{Matt~J Kusner}, \bibinfo{person}{Brooks
  Paige}, {and} \bibinfo{person}{Jos{\'e}~Miguel Hern{\'a}ndez-Lobato}.}
  \bibinfo{year}{2017}\natexlab{}.
\newblock \showarticletitle{Grammar variational autoencoder}. In
  \bibinfo{booktitle}{\emph{Proceedings of the 34th International Conference on
  Machine Learning (ICML)}}. \bibinfo{pages}{1945--1954}.
\newblock


\bibitem[\protect\citeauthoryear{Lee, Gharan, and Trevisan}{Lee
  et~al\mbox{.}}{2014}]%
        {lee2014multiway}
\bibfield{author}{\bibinfo{person}{James~R Lee}, \bibinfo{person}{Shayan~Oveis
  Gharan}, {and} \bibinfo{person}{Luca Trevisan}.}
  \bibinfo{year}{2014}\natexlab{}.
\newblock \showarticletitle{Multiway spectral partitioning and higher-order
  cheeger inequalities}.
\newblock \bibinfo{journal}{\emph{Journal of the ACM (JACM)}}
  \bibinfo{volume}{61}, \bibinfo{number}{6} (\bibinfo{year}{2014}),
  \bibinfo{pages}{1--30}.
\newblock


\bibitem[\protect\citeauthoryear{Li, Rong, Cheng, Meng, Huang, and Huang}{Li
  et~al\mbox{.}}{2019a}]%
        {jiawww19}
\bibfield{author}{\bibinfo{person}{Jia Li}, \bibinfo{person}{Yu Rong},
  \bibinfo{person}{Hong Cheng}, \bibinfo{person}{Helen Meng},
  \bibinfo{person}{Wenbing Huang}, {and} \bibinfo{person}{Junzhou Huang}.}
  \bibinfo{year}{2019}\natexlab{a}.
\newblock \showarticletitle{Semi-Supervised Graph Classification: A
  Hierarchical Graph Perspective}. In \bibinfo{booktitle}{\emph{The World Wide
  Web Conference (WWW)}}. \bibinfo{pages}{972–982}.
\newblock


\bibitem[\protect\citeauthoryear{Li, Yu, Li, Zhang, Zhao, Rong, Cheng, and
  Huang}{Li et~al\mbox{.}}{2020a}]%
        {li2020heatts}
\bibfield{author}{\bibinfo{person}{Jia Li}, \bibinfo{person}{Jianwei Yu},
  \bibinfo{person}{Jiajin Li}, \bibinfo{person}{Honglei Zhang},
  \bibinfo{person}{Kangfei Zhao}, \bibinfo{person}{Yu Rong},
  \bibinfo{person}{Hong Cheng}, {and} \bibinfo{person}{Junzhou Huang}.}
  \bibinfo{year}{2020}\natexlab{a}.
\newblock \showarticletitle{Dirichlet Graph Variational Autoencoder}. In
  \bibinfo{booktitle}{\emph{Neurips}}.
\newblock


\bibitem[\protect\citeauthoryear{Li, Zhang, Han, Rong, Cheng, and Huang}{Li
  et~al\mbox{.}}{2020b}]%
        {li2020adversarial}
\bibfield{author}{\bibinfo{person}{Jia Li}, \bibinfo{person}{Honglei Zhang},
  \bibinfo{person}{Zhichao Han}, \bibinfo{person}{Yu Rong},
  \bibinfo{person}{Hong Cheng}, {and} \bibinfo{person}{Junzhou Huang}.}
  \bibinfo{year}{2020}\natexlab{b}.
\newblock \showarticletitle{Adversarial attack on community detection by hiding
  individuals}. In \bibinfo{booktitle}{\emph{Proceedings of The Web Conference
  2020}}. \bibinfo{pages}{917--927}.
\newblock


\bibitem[\protect\citeauthoryear{Li, Zhang, Sun, Wang, Li, Zhang, and Lin}{Li
  et~al\mbox{.}}{2019b}]%
        {li2019approximate}
\bibfield{author}{\bibinfo{person}{Wen Li}, \bibinfo{person}{Ying Zhang},
  \bibinfo{person}{Yifang Sun}, \bibinfo{person}{Wei Wang},
  \bibinfo{person}{Mingjie Li}, \bibinfo{person}{Wenjie Zhang}, {and}
  \bibinfo{person}{Xuemin Lin}.} \bibinfo{year}{2019}\natexlab{b}.
\newblock \showarticletitle{Approximate nearest neighbor search on high
  dimensional data-experiments, analyses, and improvement}.
\newblock \bibinfo{journal}{\emph{IEEE Transactions on Knowledge and Data
  Engineering}} (\bibinfo{year}{2019}).
\newblock


\bibitem[\protect\citeauthoryear{Liben-Nowell and Kleinberg}{Liben-Nowell and
  Kleinberg}{2007}]%
        {liben2007link}
\bibfield{author}{\bibinfo{person}{David Liben-Nowell} {and}
  \bibinfo{person}{Jon Kleinberg}.} \bibinfo{year}{2007}\natexlab{}.
\newblock \showarticletitle{The link-prediction problem for social networks}.
\newblock \bibinfo{journal}{\emph{Journal of the American society for
  information science and technology}} \bibinfo{volume}{58},
  \bibinfo{number}{7} (\bibinfo{year}{2007}), \bibinfo{pages}{1019--1031}.
\newblock


\bibitem[\protect\citeauthoryear{Liu, Allamanis, Brockschmidt, and Gaunt}{Liu
  et~al\mbox{.}}{2018}]%
        {liu2018constrained}
\bibfield{author}{\bibinfo{person}{Qi Liu}, \bibinfo{person}{Miltiadis
  Allamanis}, \bibinfo{person}{Marc Brockschmidt}, {and}
  \bibinfo{person}{Alexander Gaunt}.} \bibinfo{year}{2018}\natexlab{}.
\newblock \showarticletitle{Constrained graph variational autoencoders for
  molecule design}. In \bibinfo{booktitle}{\emph{Proceedings of the 32nd
  International Conference on Neural Information Processing Systems
  (NeurIPS)}}. \bibinfo{pages}{7806–7815}.
\newblock


\bibitem[\protect\citeauthoryear{Lorberbom, Gane, Jaakkola, and
  Hazan}{Lorberbom et~al\mbox{.}}{2019}]%
        {lorberbom2019direct}
\bibfield{author}{\bibinfo{person}{Guy Lorberbom}, \bibinfo{person}{Andreea
  Gane}, \bibinfo{person}{Tommi Jaakkola}, {and} \bibinfo{person}{Tamir
  Hazan}.} \bibinfo{year}{2019}\natexlab{}.
\newblock \showarticletitle{Direct Optimization through $\arg\max$ for Discrete
  Variational Auto-Encoder}. In \bibinfo{booktitle}{\emph{Advances in Neural
  Information Processing Systems}}. \bibinfo{pages}{6200--6211}.
\newblock


\bibitem[\protect\citeauthoryear{Majumdar}{Majumdar}{2018}]%
        {majumdar2018blind}
\bibfield{author}{\bibinfo{person}{Angshul Majumdar}.}
  \bibinfo{year}{2018}\natexlab{}.
\newblock \showarticletitle{Blind denoising autoencoder}.
\newblock \bibinfo{journal}{\emph{IEEE transactions on neural networks and
  learning systems}} \bibinfo{volume}{30}, \bibinfo{number}{1}
  (\bibinfo{year}{2018}), \bibinfo{pages}{312--317}.
\newblock


\bibitem[\protect\citeauthoryear{Morris and Frey}{Morris and Frey}{2004}]%
        {morris2004denoising}
\bibfield{author}{\bibinfo{person}{Quaid~D Morris} {and}
  \bibinfo{person}{Brendan~J Frey}.} \bibinfo{year}{2004}\natexlab{}.
\newblock \showarticletitle{Denoising and untangling graphs using degree
  priors}. In \bibinfo{booktitle}{\emph{Proceedings of the 16th International
  Conference on Neural Information Processing Systems (NeurIPS)}}.
  \bibinfo{pages}{385--392}.
\newblock


\bibitem[\protect\citeauthoryear{Nazi, Hang, Goldie, Ravi, and Mirhoseini}{Nazi
  et~al\mbox{.}}{2019}]%
        {nazi2019gap}
\bibfield{author}{\bibinfo{person}{Azade Nazi}, \bibinfo{person}{Will Hang},
  \bibinfo{person}{Anna Goldie}, \bibinfo{person}{Sujith Ravi}, {and}
  \bibinfo{person}{Azalia Mirhoseini}.} \bibinfo{year}{2019}\natexlab{}.
\newblock \showarticletitle{GAP: Generalizable Approximate Graph Partitioning
  Framework}.
\newblock \bibinfo{journal}{\emph{International Conference on Learning
  Representations Workshop}} (\bibinfo{year}{2019}).
\newblock


\bibitem[\protect\citeauthoryear{Newman}{Newman}{2006}]%
        {newman2006modularity}
\bibfield{author}{\bibinfo{person}{Mark~EJ Newman}.}
  \bibinfo{year}{2006}\natexlab{}.
\newblock \showarticletitle{Modularity and community structure in networks}.
\newblock \bibinfo{journal}{\emph{Proceedings of the national academy of
  sciences}} \bibinfo{volume}{103}, \bibinfo{number}{23}
  (\bibinfo{year}{2006}), \bibinfo{pages}{8577--8582}.
\newblock


\bibitem[\protect\citeauthoryear{Newman}{Newman}{2009}]%
        {newman2009random}
\bibfield{author}{\bibinfo{person}{Mark~EJ Newman}.}
  \bibinfo{year}{2009}\natexlab{}.
\newblock \showarticletitle{Random graphs with clustering}.
\newblock \bibinfo{journal}{\emph{Physical review letters}}
  \bibinfo{volume}{103}, \bibinfo{number}{5} (\bibinfo{year}{2009}),
  \bibinfo{pages}{058701}.
\newblock


\bibitem[\protect\citeauthoryear{Padilla, Sharpnack, and Scott}{Padilla
  et~al\mbox{.}}{2017}]%
        {padilla2017dfs}
\bibfield{author}{\bibinfo{person}{Oscar Hernan~Madrid Padilla},
  \bibinfo{person}{James Sharpnack}, {and} \bibinfo{person}{James~G Scott}.}
  \bibinfo{year}{2017}\natexlab{}.
\newblock \showarticletitle{The DFS fused lasso: Linear-time denoising over
  general graphs}.
\newblock \bibinfo{journal}{\emph{The Journal of Machine Learning Research}}
  \bibinfo{volume}{18}, \bibinfo{number}{1} (\bibinfo{year}{2017}),
  \bibinfo{pages}{6410--6445}.
\newblock


\bibitem[\protect\citeauthoryear{Pang and Cheung}{Pang and Cheung}{2017}]%
        {pang2017graph}
\bibfield{author}{\bibinfo{person}{Jiahao Pang} {and} \bibinfo{person}{Gene
  Cheung}.} \bibinfo{year}{2017}\natexlab{}.
\newblock \showarticletitle{Graph Laplacian regularization for image denoising:
  Analysis in the continuous domain}.
\newblock \bibinfo{journal}{\emph{IEEE Transactions on Image Processing}}
  \bibinfo{volume}{26}, \bibinfo{number}{4} (\bibinfo{year}{2017}),
  \bibinfo{pages}{1770--1785}.
\newblock


\bibitem[\protect\citeauthoryear{Peng and Yoshida}{Peng and Yoshida}{2020}]%
        {peng2002aver}
\bibfield{author}{\bibinfo{person}{Pan Peng} {and} \bibinfo{person}{Yuichi
  Yoshida}.} \bibinfo{year}{2020}\natexlab{}.
\newblock \showarticletitle{Average Sensitivity of Spectral Clustering}. In
  \bibinfo{booktitle}{\emph{Proceedings of the 26th ACM SIGKDD International
  Conference on Knowledge Discovery and Data Mining (SIGKDD)}}.
\newblock


\bibitem[\protect\citeauthoryear{Richard, Savalle, and Vayatis}{Richard
  et~al\mbox{.}}{2012}]%
        {richard2012estimation}
\bibfield{author}{\bibinfo{person}{Emile Richard},
  \bibinfo{person}{Pierre-Andr{\'e} Savalle}, {and} \bibinfo{person}{Nicolas
  Vayatis}.} \bibinfo{year}{2012}\natexlab{}.
\newblock \showarticletitle{Estimation of simultaneously sparse and low rank
  matrices}. In \bibinfo{booktitle}{\emph{Proceedings of the 29th International
  Coference on International Conference on Machine Learning (ICML)}}.
  \bibinfo{pages}{51–58}.
\newblock


\bibitem[\protect\citeauthoryear{Salha, Hennequin, Remy, Moussallam, and
  Vazirgiannis}{Salha et~al\mbox{.}}{2020}]%
        {salha2020fastgae}
\bibfield{author}{\bibinfo{person}{Guillaume Salha}, \bibinfo{person}{Romain
  Hennequin}, \bibinfo{person}{Jean-Baptiste Remy}, \bibinfo{person}{Manuel
  Moussallam}, {and} \bibinfo{person}{Michalis Vazirgiannis}.}
  \bibinfo{year}{2020}\natexlab{}.
\newblock \showarticletitle{FastGAE: Fast, Scalable and Effective Graph
  Autoencoders with Stochastic Subgraph Decoding}.
\newblock \bibinfo{journal}{\emph{arXiv preprint arXiv:2002.01910}}
  (\bibinfo{year}{2020}).
\newblock


\bibitem[\protect\citeauthoryear{Salha, Hennequin, Tran, and
  Vazirgiannis}{Salha et~al\mbox{.}}{2019}]%
        {salha2019degeneracy}
\bibfield{author}{\bibinfo{person}{Guillaume Salha}, \bibinfo{person}{Romain
  Hennequin}, \bibinfo{person}{Viet~Anh Tran}, {and} \bibinfo{person}{Michalis
  Vazirgiannis}.} \bibinfo{year}{2019}\natexlab{}.
\newblock \showarticletitle{A degeneracy framework for scalable graph
  autoencoders}.
\newblock \bibinfo{journal}{\emph{Proceedings of the Twenty-Eighth
  International Joint Conference on Artificial Intelligence (IJCAI-19)}}
  (\bibinfo{year}{2019}).
\newblock


\bibitem[\protect\citeauthoryear{Samanta, Abir, Jana, Chattaraj, Ganguly, and
  Rodriguez}{Samanta et~al\mbox{.}}{2019}]%
        {samanta2019nevae}
\bibfield{author}{\bibinfo{person}{Bidisha Samanta}, \bibinfo{person}{DE Abir},
  \bibinfo{person}{Gourhari Jana}, \bibinfo{person}{Pratim~Kumar Chattaraj},
  \bibinfo{person}{Niloy Ganguly}, {and} \bibinfo{person}{Manuel~Gomez
  Rodriguez}.} \bibinfo{year}{2019}\natexlab{}.
\newblock \showarticletitle{Nevae: A deep generative model for molecular
  graphs}. In \bibinfo{booktitle}{\emph{Thirty-Third AAAI Conference on
  Artificial Intelligence (AAAI)}}. \bibinfo{pages}{1110--1117}.
\newblock


\bibitem[\protect\citeauthoryear{Sen, Namata, Bilgic, Getoor, Galligher, and
  Eliassi-Rad}{Sen et~al\mbox{.}}{2008}]%
        {sen2008collective}
\bibfield{author}{\bibinfo{person}{Prithviraj Sen}, \bibinfo{person}{Galileo
  Namata}, \bibinfo{person}{Mustafa Bilgic}, \bibinfo{person}{Lise Getoor},
  \bibinfo{person}{Brian Galligher}, {and} \bibinfo{person}{Tina Eliassi-Rad}.}
  \bibinfo{year}{2008}\natexlab{}.
\newblock \showarticletitle{Collective classification in network data}.
\newblock \bibinfo{journal}{\emph{AI magazine}} \bibinfo{volume}{29},
  \bibinfo{number}{3} (\bibinfo{year}{2008}), \bibinfo{pages}{93--106}.
\newblock


\bibitem[\protect\citeauthoryear{Shaham, Stanton, Li, Nadler, Basri, and
  Kluger}{Shaham et~al\mbox{.}}{2018}]%
        {shaham2018spectralnet}
\bibfield{author}{\bibinfo{person}{Uri Shaham}, \bibinfo{person}{Kelly
  Stanton}, \bibinfo{person}{Henry Li}, \bibinfo{person}{Boaz Nadler},
  \bibinfo{person}{Ronen Basri}, {and} \bibinfo{person}{Yuval Kluger}.}
  \bibinfo{year}{2018}\natexlab{}.
\newblock \showarticletitle{Spectralnet: Spectral clustering using deep neural
  networks}.
\newblock \bibinfo{journal}{\emph{International Conference on Learning
  Representations (ICLR)}} (\bibinfo{year}{2018}).
\newblock


\bibitem[\protect\citeauthoryear{Shervashidze, Schweitzer, Van~Leeuwen,
  Mehlhorn, and Borgwardt}{Shervashidze et~al\mbox{.}}{2011}]%
        {shervashidze2011weisfeiler}
\bibfield{author}{\bibinfo{person}{Nino Shervashidze}, \bibinfo{person}{Pascal
  Schweitzer}, \bibinfo{person}{Erik~Jan Van~Leeuwen}, \bibinfo{person}{Kurt
  Mehlhorn}, {and} \bibinfo{person}{Karsten~M Borgwardt}.}
  \bibinfo{year}{2011}\natexlab{}.
\newblock \showarticletitle{Weisfeiler-lehman graph kernels}.
\newblock \bibinfo{journal}{\emph{Journal of Machine Learning Research}}
  \bibinfo{volume}{12}, \bibinfo{number}{Sep} (\bibinfo{year}{2011}),
  \bibinfo{pages}{2539--2561}.
\newblock


\bibitem[\protect\citeauthoryear{Shi and Malik}{Shi and Malik}{2000}]%
        {shi2000normalized}
\bibfield{author}{\bibinfo{person}{Jianbo Shi} {and} \bibinfo{person}{Jitendra
  Malik}.} \bibinfo{year}{2000}\natexlab{}.
\newblock \showarticletitle{Normalized cuts and image segmentation}.
\newblock \bibinfo{journal}{\emph{IEEE Transactions on pattern analysis and
  machine intelligence}} \bibinfo{volume}{22}, \bibinfo{number}{8}
  (\bibinfo{year}{2000}), \bibinfo{pages}{888--905}.
\newblock


\bibitem[\protect\citeauthoryear{Simonovsky and Komodakis}{Simonovsky and
  Komodakis}{2018}]%
        {simonovsky2018graphvae}
\bibfield{author}{\bibinfo{person}{Martin Simonovsky} {and}
  \bibinfo{person}{Nikos Komodakis}.} \bibinfo{year}{2018}\natexlab{}.
\newblock \showarticletitle{Graphvae: Towards generation of small graphs using
  variational autoencoders}. In \bibinfo{booktitle}{\emph{Artificial Neural
  Networks and Machine Learning (ICANN)}}. Springer, \bibinfo{pages}{412--422}.
\newblock


\bibitem[\protect\citeauthoryear{Varma and Yoshida}{Varma and Yoshida}{2019}]%
        {varma2019average}
\bibfield{author}{\bibinfo{person}{Nithin Varma} {and} \bibinfo{person}{Yuichi
  Yoshida}.} \bibinfo{year}{2019}\natexlab{}.
\newblock \showarticletitle{Average sensitivity of graph algorithms}.
\newblock \bibinfo{journal}{\emph{arXiv preprint arXiv:1904.03248}}
  (\bibinfo{year}{2019}).
\newblock


\bibitem[\protect\citeauthoryear{Von~Luxburg}{Von~Luxburg}{2007}]%
        {von2007tutorial}
\bibfield{author}{\bibinfo{person}{Ulrike Von~Luxburg}.}
  \bibinfo{year}{2007}\natexlab{}.
\newblock \showarticletitle{A tutorial on spectral clustering}.
\newblock \bibinfo{journal}{\emph{Statistics and computing}}
  \bibinfo{volume}{17}, \bibinfo{number}{4} (\bibinfo{year}{2007}),
  \bibinfo{pages}{395--416}.
\newblock


\bibitem[\protect\citeauthoryear{Wang, Pourshafeie, Zitnik, Zhu, Bustamante,
  Batzoglou, and Leskovec}{Wang et~al\mbox{.}}{2018}]%
        {wang2018network}
\bibfield{author}{\bibinfo{person}{Bo Wang}, \bibinfo{person}{Armin
  Pourshafeie}, \bibinfo{person}{Marinka Zitnik}, \bibinfo{person}{Junjie Zhu},
  \bibinfo{person}{Carlos~D Bustamante}, \bibinfo{person}{Serafim Batzoglou},
  {and} \bibinfo{person}{Jure Leskovec}.} \bibinfo{year}{2018}\natexlab{}.
\newblock \showarticletitle{Network enhancement as a general method to denoise
  weighted biological networks}.
\newblock \bibinfo{journal}{\emph{Nature communications}} \bibinfo{volume}{9},
  \bibinfo{number}{1} (\bibinfo{year}{2018}), \bibinfo{pages}{1--8}.
\newblock


\bibitem[\protect\citeauthoryear{Wang, Wang, Yu, and Zhang}{Wang
  et~al\mbox{.}}{2015}]%
        {wang2015community}
\bibfield{author}{\bibinfo{person}{Meng Wang}, \bibinfo{person}{Chaokun Wang},
  \bibinfo{person}{Jeffrey~Xu Yu}, {and} \bibinfo{person}{Jun Zhang}.}
  \bibinfo{year}{2015}\natexlab{}.
\newblock \showarticletitle{Community detection in social networks: an in-depth
  benchmarking study with a procedure-oriented framework}.
\newblock \bibinfo{journal}{\emph{VLDB}} \bibinfo{volume}{8},
  \bibinfo{number}{10} (\bibinfo{year}{2015}), \bibinfo{pages}{998--1009}.
\newblock


\bibitem[\protect\citeauthoryear{Wu, Souza, Zhang, Fifty, Yu, and
  Weinberger}{Wu et~al\mbox{.}}{2019}]%
        {pmlr-v97-wu19e}
\bibfield{author}{\bibinfo{person}{Felix Wu}, \bibinfo{person}{Amauri Souza},
  \bibinfo{person}{Tianyi Zhang}, \bibinfo{person}{Christopher Fifty},
  \bibinfo{person}{Tao Yu}, {and} \bibinfo{person}{Kilian Weinberger}.}
  \bibinfo{year}{2019}\natexlab{}.
\newblock \showarticletitle{Simplifying Graph Convolutional Networks}. In
  \bibinfo{booktitle}{\emph{Proceedings of the 36th International Conference on
  Machine Learning (ICML)}}. \bibinfo{publisher}{PMLR},
  \bibinfo{pages}{6861--6871}.
\newblock


\bibitem[\protect\citeauthoryear{{Xu}, {Yang}, {Wang}, {Liu}, {Zhang}, {Chen},
  and {Lu}}{{Xu} et~al\mbox{.}}{2020}]%
        {9200731}
\bibfield{author}{\bibinfo{person}{J. {Xu}}, \bibinfo{person}{Y. {Yang}},
  \bibinfo{person}{C. {Wang}}, \bibinfo{person}{Z. {Liu}}, \bibinfo{person}{J.
  {Zhang}}, \bibinfo{person}{L. {Chen}}, {and} \bibinfo{person}{J. {Lu}}.}
  \bibinfo{year}{2020}\natexlab{}.
\newblock \showarticletitle{Robust Network Enhancement from Flawed Networks}.
\newblock \bibinfo{journal}{\emph{IEEE Transactions on Knowledge and Data
  Engineering}} (\bibinfo{year}{2020}), \bibinfo{pages}{1--1}.
\newblock


\bibitem[\protect\citeauthoryear{Yanardag and Vishwanathan}{Yanardag and
  Vishwanathan}{2015}]%
        {Yanardag:2015}
\bibfield{author}{\bibinfo{person}{P. Yanardag} {and} \bibinfo{person}{S.V.N.
  Vishwanathan}.} \bibinfo{year}{2015}\natexlab{}.
\newblock \showarticletitle{Deep Graph Kernels}. In
  \bibinfo{booktitle}{\emph{KDD}}. \bibinfo{pages}{1365--1374}.
\newblock


\bibitem[\protect\citeauthoryear{Yang, Liu, Zhao, Sun, and Chang}{Yang
  et~al\mbox{.}}{2015}]%
        {yang2015network}
\bibfield{author}{\bibinfo{person}{Cheng Yang}, \bibinfo{person}{Zhiyuan Liu},
  \bibinfo{person}{Deli Zhao}, \bibinfo{person}{Maosong Sun}, {and}
  \bibinfo{person}{Edward Chang}.} \bibinfo{year}{2015}\natexlab{}.
\newblock \showarticletitle{Network representation learning with rich text
  information}. In \bibinfo{booktitle}{\emph{The International Joint Conference
  on Artificial Intelligence (IJCAL)}}.
\newblock


\bibitem[\protect\citeauthoryear{Zeng, Pang, Sun, and Cheung}{Zeng
  et~al\mbox{.}}{2019}]%
        {zeng2019deep}
\bibfield{author}{\bibinfo{person}{Jin Zeng}, \bibinfo{person}{Jiahao Pang},
  \bibinfo{person}{Wenxiu Sun}, {and} \bibinfo{person}{Gene Cheung}.}
  \bibinfo{year}{2019}\natexlab{}.
\newblock \showarticletitle{Deep graph Laplacian regularization for robust
  denoising of real images}. In \bibinfo{booktitle}{\emph{Proceedings of the
  IEEE Conference on Computer Vision and Pattern Recognition (CVPR) Workshop}}.
\newblock


\bibitem[\protect\citeauthoryear{Zhang, Yao, Huang, Jiang, Li, and
  Chawla}{Zhang et~al\mbox{.}}{2020}]%
        {zhang2019few}
\bibfield{author}{\bibinfo{person}{Chuxu Zhang}, \bibinfo{person}{Huaxiu Yao},
  \bibinfo{person}{Chao Huang}, \bibinfo{person}{Meng Jiang},
  \bibinfo{person}{Zhenhui Li}, {and} \bibinfo{person}{Nitesh~V Chawla}.}
  \bibinfo{year}{2020}\natexlab{}.
\newblock \showarticletitle{Few-Shot Knowledge Graph Completion}.
\newblock \bibinfo{journal}{\emph{AAAI}} (\bibinfo{year}{2020}).
\newblock


\bibitem[\protect\citeauthoryear{Zhang and Chen}{Zhang and Chen}{2018}]%
        {zhang2018link}
\bibfield{author}{\bibinfo{person}{Muhan Zhang} {and} \bibinfo{person}{Yixin
  Chen}.} \bibinfo{year}{2018}\natexlab{}.
\newblock \showarticletitle{Link prediction based on graph neural networks}. In
  \bibinfo{booktitle}{\emph{Proceedings of the 32nd International Conference on
  Neural Information Processing Systems (NeurIPS)}}.
  \bibinfo{pages}{5171–5181}.
\newblock


\bibitem[\protect\citeauthoryear{Zhou, Cheng, and Yu}{Zhou
  et~al\mbox{.}}{2009}]%
        {zhou2009graph}
\bibfield{author}{\bibinfo{person}{Yang Zhou}, \bibinfo{person}{Hong Cheng},
  {and} \bibinfo{person}{Jeffrey~Xu Yu}.} \bibinfo{year}{2009}\natexlab{}.
\newblock \showarticletitle{Graph clustering based on structural/attribute
  similarities}.
\newblock \bibinfo{journal}{\emph{Proceedings of the VLDB Endowment}}
  \bibinfo{volume}{2}, \bibinfo{number}{1} (\bibinfo{year}{2009}),
  \bibinfo{pages}{718--729}.
\newblock


\bibitem[\protect\citeauthoryear{Z{\"u}gner, Akbarnejad, and
  G{\"u}nnemann}{Z{\"u}gner et~al\mbox{.}}{2018}]%
        {zugner2018adversarial}
\bibfield{author}{\bibinfo{person}{Daniel Z{\"u}gner}, \bibinfo{person}{Amir
  Akbarnejad}, {and} \bibinfo{person}{Stephan G{\"u}nnemann}.}
  \bibinfo{year}{2018}\natexlab{}.
\newblock \showarticletitle{Adversarial Attacks on Neural Networks for Graph
  Data}. In \bibinfo{booktitle}{\emph{KDD}}. \bibinfo{pages}{2847--2856}.
\newblock


\end{thebibliography}
